%% file: main.tex
\ifdefined\XeTeXversion\else\pdfoutput=1\fi
\documentclass[10pt,twocolumn,letterpaper]{article}

\usepackage[pagenumbers]{cvpr}

\input{preamble}

\definecolor{cvprblue}{rgb}{0.21,0.49,0.74}
\usepackage[pagebackref,breaklinks,colorlinks,allcolors=cvprblue]{hyperref}

\title{Does the Model Use What It Cites?\\Counterfactual Tests of Grounded Reasoning Chains in Small Video--Language Models}

\author{Jugal Gajjar\textsuperscript{*} \qquad Kamalasankari Subramaniakuppusamy\textsuperscript{*}\\
Independent Researchers\\
{\tt\small 812jugalgajjar@gmail.com \qquad iamkamalamskls@gmail.com}\\[2pt]
{\small Code: \href{https://github.com/JugalGajjar/CausalSight}{\texttt{github.com/JugalGajjar/CausalSight}}}}

\begin{document}
\maketitle
\renewcommand{\thefootnote}{\fnsymbol{footnote}}\footnotetext[1]{Equal contribution.}\renewcommand{\thefootnote}{\arabic{footnote}}
\input{sec/0_abstract}
\input{sec/1_intro}
\input{sec/2_related}
\input{sec/3_method}
\input{sec/4_faithfulness}
\input{sec/5_experiments}
\input{sec/6_discussion}
\input{sec/7_conclusion}
{
    \small
    \bibliographystyle{ieeenat_fullname}
    \bibliography{main}
}

\input{sec/X_suppl}

\end{document}

%% file: preamble.tex
\usepackage[expansion=false]{microtype}
\usepackage[T1]{fontenc}
\usepackage{tikz}
\usetikzlibrary{arrows.meta,positioning,calc}
\usepackage{multirow}
\usepackage{xspace}

\newcommand{\citegap}{cite gap\xspace}
\newcommand{\CI}[2]{{\scriptsize[#1,\,#2]}}
\newcommand{\tabsize}{\footnotesize}

%% file: sec/0_abstract.tex
\begin{abstract}
A grounded reasoning chain cites a frame span and a box for every step, so a reader can check where the model looked. Whether the answer \emph{depends} on the cited region is a different question. We test it by intervention: the regions a model cited in its own chain are blacked out, and the share of correct answers that flip is compared with blacking out boxes of the same size elsewhere on the same frames. The paired difference, the \citegap, needs no annotation. We apply it to 2--3B video--language models trained on exact CLEVRER chains and on pseudo-labelled NExT-QA chains. With the same weights, the \citegap is 0.089 on CLEVRER and 0.003 on NExT-QA. Pooled over models, masking the cited boxes flips 2.2 times as many correct answers as the control on CLEVRER and 1.1 times as many on NExT-QA, also among the answers that depend on the video at all, while the boxes pass a grounding check on both. Masking the cited \emph{entity} on every frame, instead of the cited box, flips 1.9 times as many answers as its control. The NExT-QA chains were supervised to cite two frames of entities that stay visible on most frames, so this outcome is what the supervision predicts; the point is that the usual check on grounded chains cannot see it. A citation can be verifiable and relevant without the answer depending on it, and a test of grounded reasoning should remove what a chain names, not only where it points.
\end{abstract}

%% file: sec/1_intro.tex
\section{Introduction}
\label{sec:intro}

Reinforcement learning with verifiable rewards has become the standard way to make multimodal models reason~\cite{shao2024deepseekmath,deepseekr1,feng2025videor1,wang2025videorft}. For video, where the evidence for an answer is spread over time and space, the natural refinement is a \emph{grounded} chain: each step names what the model looked at, where, and when~\cite{shao2024visualcot,cao2025groundr1,peh2026hgrpo}. Such a chain promises two things. It is \emph{verifiable}: a reader can check whether the cited region contains what the step claims. And it is \emph{faithful}: the answer is supposedly computed from the cited evidence. The first promise can be checked with a detector. The second is a claim about the model's computation, and it is the one that matters if grounded chains are to serve as explanations. It has a consequence that can be tested at the input: if the answer rests on the cited evidence, removing that evidence should change the answer more than removing something else.

We test this consequence on small video--language models (VLMs), the 2--3B class, for two reasons. It is the class that is cheap enough to fine-tune, run on device, and evaluate under a suite of full-pass interventions per model. It is the class for which RL-induced shortcut behaviour has been reported~\cite{wu2026beyond}, so the question of what its reasoning rests on is open. We train Qwen2.5-VL-3B~\cite{bai2025qwen25vl} and InternVL3-2B~\cite{zhu2025internvl3} to answer causal video questions through chains of grounded triplets (sub-question, intermediate answer, spatiotemporal evidence) and ask whether the answers depend on what the chains cite.

\begin{figure}[t]
\centering
\includegraphics[width=\linewidth]{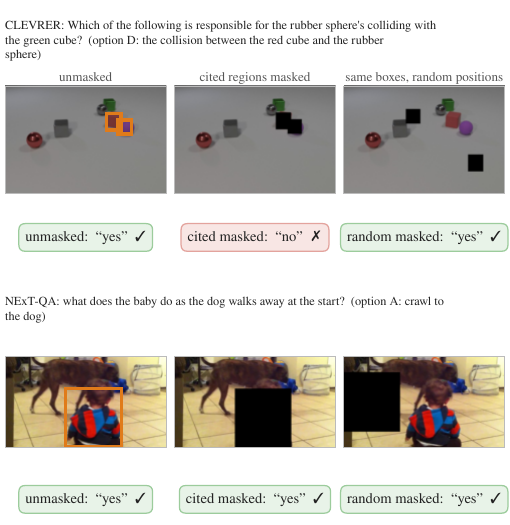}
\caption{\textbf{The self-citation test on two items from the same model.} Left: the frame with the regions the mixed chain SFT model cited (orange). Middle: those regions blacked out. Right: boxes of the same size at random positions (illustrative placement). On CLEVRER (top) masking the citations flips the answer and masking elsewhere does not; on NExT-QA (bottom) the citations sit on the named entities and the answer survives either mask.}
\label{fig:teaser}
\end{figure}

\paragraph{A counterfactual self-citation test.}
The tool is simple (\cref{fig:teaser}). Run the model once and record the regions it cited. Run it again with exactly those regions blacked out. Run it a third time with boxes of the same sizes, on the same frames, moved to random positions. The fraction of correct answers that flip under the first intervention, minus the fraction that flip under the second, is the \emph{\citegap}: how much more the answer depends on the cited pixels than on an equal area of occlusion elsewhere. Both interventions remove the same area at the same times, so the gap cannot be explained by occlusion itself; and the regions come from the model's own output, so the test applies to any video, annotated or not. Three controls from the same family accompany it: a blind gap (all frames black), a gold-evidence gap (annotated evidence against size-matched random boxes) and an object-track gap (the decisive object removed for the whole video against size-matched copies elsewhere). The masking gaps are paired differences in flip rates with bootstrap intervals.

\paragraph{Findings.}
The same weights behave differently on the two benchmarks. For the model trained on a mix of exact CLEVRER~\cite{yi2020clevrer} chains and pseudo-labelled NExT-QA~\cite{xiao2021nextqa} chains, the \citegap is 0.089 on CLEVRER and 0.003 on NExT-QA; for the three other models tested on both it is 0.051 against 0.015, 0.042 against 0.004 and 0.030 against 0.001. Pooled over five chain models, blacking out the cited boxes flips 9.7\% of correct answers against 4.5\% for the control on CLEVRER, and 2.3\% against 2.1\% on NExT-QA. The NExT-QA gap stays small among the answers that depend on the video, on correct options, where every cited box matches a detector~\cite{liu2024groundingdino}, and where the control leaves the cited area untouched (\cref{sec:results_real}). The cited \emph{entity} matters more than the cited box. It is visible on 14 of 16 frames while a citation covers two, and blacking it out on all of them flips 10.2\% of correct answers against 5.5\%; among questions answered correctly, the selected answer changes for 35\% against 20\%. Given that supervision, a small \citegap is unsurprising. What the case shows is that the usual check does not reveal it: the NExT-QA boxes agree with a detector, as the CLEVRER boxes agree with the annotation.

\paragraph{Contributions.}
(i) Two annotation-free tests for grounded reasoning chains: a counterfactual self-citation test with a size-matched control and question-clustered paired intervals, and an entity-track test that removes what a chain names rather than where it points; with them, the diagnostics needed to read a gap (the intervention's own flip rate as a bound, video dependence, control overlap and object-level controls, citation placement).
(ii) A case study on 2--3B VLMs trained to reason in grounded chains: exact CLEVRER chains from executed programs (100\% agreement with ground truth over 256k checks), pseudo-labelled NExT-QA chains with a manual audit, supervised fine-tuning and GRPO on chains.
(iii) The finding that detector-agreement grounding does not distinguish citations the answer depends on from citations it does not: the boxes pass a grounding check on both benchmarks.

%% file: sec/2_related.tex
\section{Related Work}
\label{sec:related}

\paragraph{RL for video reasoning in small VLMs.}
GRPO~\cite{shao2024deepseekmath,deepseekr1} with outcome and format rewards is the standard recipe for video reasoning models~\cite{feng2025videor1,wang2025videorft,zhang2025tinyllavavideor1,wang2025clgrpo}. Wu~\etal~\cite{wu2026beyond} find that outcome-only GRPO on Qwen2.5-VL-3B lowers CLEVRER inferential accuracy by 10 points and blame perceptual shortcuts; Visionary-R1~\cite{xia2025visionaryr1} and COPO~\cite{guo2026copo} target shortcuts and hallucination under RL. We did not observe the collapse under LoRA (\cref{sec:supp_collapse}) and evaluate Wu~\etal's checkpoint under our suite.

\paragraph{Grounded chains of thought.}
Visual CoT~\cite{shao2024visualcot}, DeepEyes~\cite{zheng2026deepeyes}, Ground-R1~\cite{cao2025groundr1}, VLM-R$^3$~\cite{jiang2025vlmr3} and Chain-of-Focus~\cite{zhang2025chainoffocus} make image reasoning emit or attend to regions. For video, Open-o3-Video~\cite{meng2025openo3video} emits timestamps and boxes alongside the answer, H-GRPO~\cite{peh2026hgrpo} rewards a set of grounded triplets through permutation-invariant Hungarian matching on image-only evidence, and CFPO~\cite{yu2026cfpo} uses counterfactual rollouts as a training signal. Our chains extend the triplet to video with spans, boxes and dependency links; our process rewards, which reuse H-GRPO's matching, add nothing measurable at 3B (\cref{sec:supp_stage2}). In this work grounding is scored by \emph{where} a region is, against a reference box or a detector, not by whether the answer needs the region.

\begin{figure*}[t]
\centering
\includegraphics[width=\textwidth]{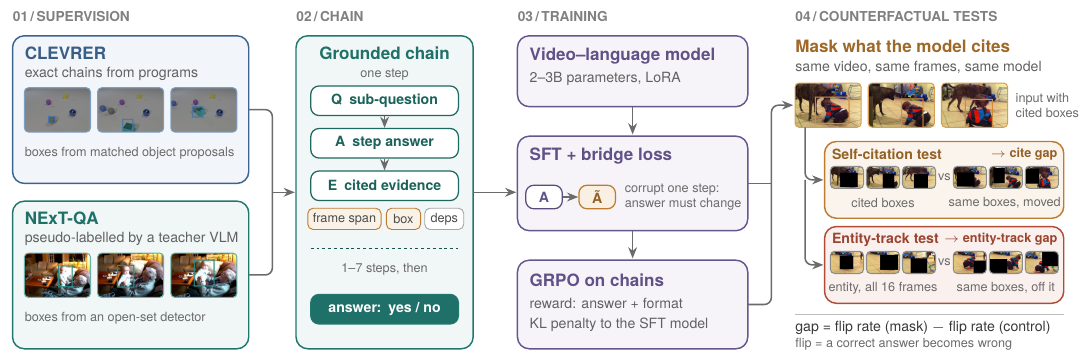}
\caption{\textbf{Pipeline and tests.} Chains are generated exactly for CLEVRER and pseudo-labelled for NExT-QA (real frames with the evidence boxes of one chain), and a 2--3B model learns to emit them. Both tests start from the regions the trained model cited: the self-citation test masks the cited boxes or the same boxes moved elsewhere; the entity-track test does the same for the named entity on every frame.}
\label{fig:pipeline}
\end{figure*}

\paragraph{Faithfulness of explanations.}
Faithfulness is the claim that an explanation reflects the computation behind the answer~\cite{jacovi2020faithfully}; for chains of thought it has been tested by perturbing the chain~\cite{lanham2023faithfulness,turpin2023unfaithful,chen2025reasoningmodels,parcalabescu2024faithfulness}. RSAT~\cite{gajjar2026rsat} uses reinforcement learning to make small language models attribute table reasoning to cells faithfully, and citations of retrieved passages can be correct yet post-rationalised~\cite{wallat2024correctness}; our attributed unit is a spatiotemporal region, and our check is an intervention rather than agreement with a reference. In vision, attention faithfulness is heterogeneous across models~\cite{song2026attention}, VLMs answer from priors~\cite{vo2025biased}, and VQA debiasing has long used counterfactual and ensemble objectives~\cite{agrawal2018dont,cadene2019rubi,clark2019dont,niu2021cfvqa}; grounding supervision from human attention~\cite{selvaraju2019hint,wu2019selfcritical} and from constrained explanations~\cite{ross2017right} trains a model to look at the right place, which is again a claim about location; for VQA such gains have been shown not to come from the grounding itself~\cite{shrestha2020negative}, and FPVG~\cite{reich2023fpvg} contrasts removing annotated relevant and irrelevant objects. Our test perturbs the \emph{input} where the model itself cited, in the tradition of occlusion and erasure analyses~\cite{zeiler2014visualizing,petsiuk2018rise,hooker2019roar}, turning a claim about the explanation into an intervention on the evidence. In the terms of ERASER~\cite{deyoung2020eraser} it is a comprehensiveness test with a size-matched random-rationale baseline, applied to citations the model itself emits in video; we do not test sufficiency. Masks are themselves out of distribution~\cite{fong2017meaningful,hase2021ood}, which is why every gap is taken against a matched mask. Whether an explanation is \emph{stable} under input perturbation is a separate property, to be measured only where the prediction is preserved~\cite{subramaniakuppusamy2026feature}.

\paragraph{Benchmarks.}
CLEVRER~\cite{yi2020clevrer} provides programs and object annotations, which let us build chains and gold-evidence controls exactly; NExT-QA~\cite{xiao2021nextqa} and CausalVQA~\cite{foss2025causalvqa} are real-video causal benchmarks without spatial evidence annotation. Closest in spirit, NExT-GQA~\cite{xiao2024nextgqa} adds temporal spans to NExT-QA and finds that models answer correctly while grounding the wrong moment; our test asks the same about the model's own spatial citations and needs no span labels. That NExT-QA and similar benchmarks can often be answered from a single well-chosen frame~\cite{buch2022revisiting,lei2023revealing} is the redundancy that our entity-track test addresses, and invariant grounding~\cite{li2022invariant} trains video QA models to rely on a causal scene rather than its complement.

%% file: sec/3_method.tex
\section{Grounded Chains for Small VLMs}
\label{sec:method}

\cref{fig:pipeline} gives the overview.

\subsection{Chain format}
A chain is written inside \texttt{<think>} tags, one triplet per line, followed by the answer:
\begin{center}\scriptsize\ttfamily
[n] Q: <sub-question> | A: <intermediate answer> |\\
\phantom{[n]} E: t=<start>-<end> box=(x0,y0,x1,y1) | deps=<ids>
\end{center}
The span $t$ indexes a 128-position time grid over the clip, the box is in normalised image coordinates, and \texttt{deps} lists the earlier steps the step depends on. One parser serves the format reward, the process rewards and the self-citation test, so a citation is exactly what the model wrote in an \texttt{E:} field.

\subsection{Chain supervision}
\paragraph{CLEVRER.} CLEVRER questions come with executable programs. We run each program through a traced executor over the ground-truth scene and turn every entity-producing step into a triplet: the evidence box comes from the released object proposals, matched to the ground-truth object by its unique attributes; the span is the interval over which the step's object or event is observed; dependency links follow the program's data flow. Predictive and counterfactual outcomes require the physics simulator, so their final answers come from the choice labels, but every observable intermediate step is grounded. After calibrating three annotation conventions (\cref{sec:supp_training}), the executor reproduces the ground truth on all 256,449 descriptive questions and explanatory choices of the training and validation splits. Chains whose evidence box cannot be recovered (3.5\%) are dropped. Each (question, choice) pair yields one chain and one yes/no item.

\paragraph{NExT-QA.} NExT-QA has no evidence annotation, so its chains are pseudo-labelled. A teacher (Qwen2.5-VL-7B) sees 16 frames, the question, the options and the correct option, and writes two to four observation steps, each naming a visible entity and the frame where it appears. Grounding DINO~\cite{liu2024groundingdino} (tiny, threshold 0.35) localises the entity on that frame; undetected steps are dropped, and so are chains with no surviving step. The detector box becomes the step's evidence and the frame index becomes a span of $\pm4$ grid positions. Each chain yields two training items with the same steps, the correct option answered ``yes'' and one sampled wrong option answered ``no'', so the steps do not depend on the option. In a manual audit of 100 chains (\cref{tab:audit}, supplement) the box is on the named entity in 70\% of the 299 steps and 95\% of steps are relevant, but 92 chains contain only observations (``Where is the white dog?'') and no causal sub-question (\cref{sec:discussion}).

\subsection{Training}
\paragraph{Supervised fine-tuning with a bridge loss.} We fine-tune with LoRA~\cite{hu2022lora} (rank 64 on all attention and MLP projections) to emit the chain and the answer given the frames and the question. The token-level loss alone does not make the answer depend on the chain, so for half of the examples we add a bridge term. Let $c$ be the reference chain and $\tilde c$ the same chain with the intermediate answer of one step on the dependency path to the final answer corrupted (an attribute swapped, a yes/no flipped, or a frame index shifted). With $a^\ast$ the correct answer,
\begin{equation}
\mathcal{L} = -\log p(c, a^\ast \mid v, q) \;-\; \lambda\, \log\big(1 - p(a^\ast \mid v, q, \tilde c)\big),
\label{eq:bridge}
\end{equation}
which penalises producing the correct answer from a corrupted chain. We use $\lambda=0.2$ on 50\% of examples.

\paragraph{GRPO on chains.} From the SFT checkpoint we run GRPO~\cite{shao2024deepseekmath}: for each prompt a group of $G=8$ complete chains is sampled at temperature 1, scored with a reward $R$, and updated with the clipped, group-normalised objective and a KL penalty ($\beta=0.04$) to the SFT model. The reward is $R = R_{\text{out}} + 0.5\,R_{\text{fmt}}$, with $R_{\text{out}}\in\{0,1\}$ answer correctness and $R_{\text{fmt}}\in\{0,1\}$ whether the chain parses. We also trained arms with grounding, step-necessity and order-aware process rewards and with step-level credit assignment; none moved accuracy or grounding beyond seed noise, and they are described and reported in the supplement (\cref{sec:supp_stage2}).

\paragraph{Data mixes.} The \emph{CLEVRER chain} pipeline uses 8,800 CLEVRER training chains for SFT and 1,500 CLEVRER per-option items (375 per type) for RL. The \emph{mixed} pipeline uses 4,400 CLEVRER chains plus 8,000 NExT-QA chain items (4,000 ``yes'' and 4,000 ``no'', a balanced sample from the chains of 5,283 questions) for SFT and 1,500 per-option items (750 per source) for RL; the \emph{NExT-QA chain} pipeline uses the NExT-QA items alone. The mixed arm has a \emph{plain} (chain-free) counterpart trained on the same questions with a bare \texttt{<answer>}, which separates the effect of the chains from the effect of the data; for the CLEVRER chain arm the chain-free reference is outcome-only GRPO from the base. Hyper-parameters are in the supplement (\cref{sec:supp_training}).

%% file: sec/4_faithfulness.tex
\section{Measuring Whether Citations Are Needed}
\label{sec:faithfulness}

Every masking test in the suite has the same form: an \emph{intervention} on the input frames, a \emph{size-matched control}, and their paired difference over items. Let $y_i^{\text{plain}}$, $y_i^{\text{int}}$ and $y_i^{\text{ctl}}$ be a model's answers to item $i$ under the unmodified input, the intervention and the control, and let $S=\{i: y_i^{\text{plain}} \text{ correct}\}$. The flip rate of a condition is the share of $S$ it breaks, and the gap is the paired difference,
\begin{equation}
f_{\text{int}} = \tfrac{1}{|S|}\textstyle\sum_{i\in S}\mathbf{1}[y_i^{\text{int}}\ \text{wrong}],\qquad
\Delta = f_{\text{int}} - f_{\text{ctl}},
\label{eq:gap}
\end{equation}
with a 95\% interval from a paired bootstrap (2,000 resamples) over \emph{questions}, so the per-option items of one question are resampled together and their correlation is not mistaken for independent evidence. Because the control removes the same area at the same times, $\Delta$ reads as ``how much more the answer depends on these pixels than on an equal area of other pixels''. Items undefined under one condition are dropped from both, so the two flip rates cover the same items. We read a gap by its size and by the ratio $f_{\text{int}}/f_{\text{ctl}}$, not as a binary verdict, and we report $f_{\text{int}}$ itself: the gap cannot exceed it under any control, so it bounds what the intervened pixels can account for. Masked area is given as a share of \emph{pixel-frames}, the pixels of the 16 sampled frames.

\paragraph{Blind gap.} Plain accuracy minus accuracy with every frame black. A model that answers from the question alone scores zero; the gap is a floor on how much the model uses the video at all.

\paragraph{Gold-evidence gap (CLEVRER only).} The annotated evidence regions of the question's chain are blacked out on the sampled frames nearest their spans; the control blacks out boxes of the same sizes and spans at random positions.

\paragraph{Object-track gap (CLEVRER only).} On CLEVRER a single-frame box removes little, because the object is visible in the other frames. We therefore remove the decisive evidence object(s) of the chain (the candidate collision of the option for predictive and counterfactual questions, otherwise the last step) for the whole video, against size-matched copies of the same tracks at random positions that avoid the evidence. This is the strongest annotated test and applies to chain-free models too.

\paragraph{Cite gap (the self-citation test).} The intervention blacks out the regions cited in the model's \emph{own} chain on the unmodified input (every \texttt{E:} field; spans mapped from the 128-grid onto the 16 sampled frames; boxes covering the whole frame dropped). The control blacks out boxes of the same size on the same frames at uniformly random positions. It may overlap a cited box, which shrinks the gap: conservative when the gap is positive, the opposite when it is near zero. We therefore also report the gap on the items whose control leaves the cited area untouched (the control is a seeded draw and is replayed exactly). Under both interventions the model runs afresh, writing a new chain; only the answer is scored. Items whose chain cites no localisable region are dropped. The regions come from the model under test, so the test needs no annotation. It is a test of necessity at the input, not of whether the answer is computed through the chain (\cref{sec:discussion}). A final control checks the masking itself: a chain-free model trained on the same questions is masked with a chain model's citations and should show $\Delta\approx0$.

\paragraph{Entity-track gap (NExT-QA).} The object-track gap for video without annotation, with the chain supplying the object and a detector the track: every entity named by a ``Where is $\langle$entity$\rangle$?'' step is re-detected with Grounding DINO on all 16 sampled frames, every detection plus the cited boxes is blacked out, and the control places boxes of the same sizes on the same frames at random positions that avoid the entity boxes. Items whose entities are never detected are dropped. The chain names the entities and the detector finds them, so this test needs no annotation either.

\paragraph{Citation grounding.} The \citegap says whether the answer depends on a citation, not whether the citation is correct. On NExT-QA we check correctness separately: for every step of the form ``Where is $\langle$entity$\rangle$?'' the entity is re-detected with Grounding DINO on the cited frame, and the cited box is scored against the top detection (IoU, hit at IoU 0.5) alongside a same-size box at a random position. The detector is the family that produced the training boxes, so this measures agreement with the supervision, not ground truth.

\paragraph{Per-option evaluation.} Chain models answer one yes/no item per option, as trained, so a five-option question becomes five items. We report per-option accuracy (always-``no'' baseline 0.80 on NExT-QA) and a \emph{selection} accuracy that picks the option judged ``yes'' (credit split on ties, $1/5$ if none), which is comparable to a \emph{letters prompt} (one multiple-choice prompt answered with an option letter). Untrained reference models are evaluated in the same format, because the format itself costs accuracy (\cref{sec:results_real}). Four of five NExT-QA items are wrong options, and a correct rejection rarely flips under a box mask, so pooled flip rates are diluted; we report the NExT-QA gaps also on the items whose option is correct.

%% file: sec/5_experiments.tex
\section{Experiments}
\label{sec:experiments}

\subsection{Setup}
\paragraph{Models.} Qwen2.5-VL-3B-Instruct~\cite{bai2025qwen25vl} is the main base and InternVL3-2B~\cite{zhu2025internvl3} the second, used for the synthetic pipeline. Every model sees 16 uniformly sampled frames at native resolution, decodes greedily, and is evaluated with identical preprocessing.

\paragraph{Benchmarks.} CLEVRER validation: 1,000 stratified questions (250 per type), which become 2,537 per-option items (250 open-ended descriptive questions scored per question, 2,287 options scored yes/no). NExT-QA validation: 999 questions (333 causal, temporal and descriptive), 4,995 per-option items. Every condition of the suite runs on the same items. CausalVQA~\cite{foss2025causalvqa} is in the supplement.

\paragraph{Baselines.} Zero-shot bases; outcome-only GRPO on CLEVRER; the released VideoRFT-3B~\cite{wang2025videorft}, VideoThinker-R1-3B~\cite{wu2026beyond} and Video-R1-7B~\cite{feng2025videor1} checkpoints; and the plain (chain-free) counterpart of the mixed arm.

\begin{table*}[t]
\centering\tabsize\setlength{\tabcolsep}{5pt}
\caption{\textbf{Self-citation test on CLEVRER} (2,537 per-option items). $n$ = items whose chain cites a localisable region; flip = share of the correct ones among them that become wrong when the model's own cited regions (cited) or same-size boxes at random positions on the same frames (random) are blacked out. The \citegap is their paired difference and the gold-evidence gap the same statistic for the annotated evidence; 95\% intervals in brackets. Hit@0.5 is the share of cited boxes with IoU $\ge0.5$ against the named object. The CLEVRER chain GRPO row is seed 0.}
\label{tab:cite_clevrer}
\input{tables/tab_cite_clevrer}
\end{table*}

\paragraph{Accuracy and grounding on CLEVRER.}\label{sec:results_clevrer} The CLEVRER chain SFT model reaches 0.851 per option with 99\% parseable chains and a box IoU of 0.68 against the named object; chain GRPO adds 1.9 points (0.870\,$\pm$\,0.001 over three seeds) at IoU 0.66 (\cref{tab:clevrer_main}, supplement). Against outcome-only GRPO, which writes no chain, the chain models are higher on descriptive questions, level on explanatory and lower on predictive and counterfactual ones, and mixing chain styles costs more: 0.739 per option for the mixed chain SFT against 0.906 for its plain counterpart. Two side results are in the supplement: outcome-only GRPO under LoRA does not reproduce the shortcut collapse reported for this base~\cite{wu2026beyond}, and process rewards have no measurable effect over two to three seeds.

\subsection{CLEVRER: answers depend on the cited box}
\label{sec:results_cite_clevrer}

\cref{tab:cite_clevrer} and \cref{fig:cite_gap} (left) apply the self-citation test to five chain models on CLEVRER, one per training recipe and base. All five gaps are positive with intervals above zero, from 0.030 for the model trained on NExT-QA chains only to 0.089 for the mixed chain SFT. Pooled over the five models, blacking out the cited boxes flips 9.7\% of correct answers against 4.5\% for the control: a gap of 0.053 \CI{0.045}{0.060} and a ratio of 2.2 (1.8 to 2.4 per model). The \citegap is below the gold-evidence gap of the same model in every row (0.062--0.094), as expected when citations overlap the annotated evidence only partly. It is positive on correct and on wrong options for every model (0.02--0.09 and 0.03--0.11; \cref{tab:gold_split}, supplement), so the cited evidence moves the answer in both directions; net accuracy falls by 0.04--0.07 under the cited mask against 0.01--0.02 under the control, for the models trained with CLEVRER chains. The control barely touches the cited boxes here (6--10\% of their area).

Two cautions apply. Chain-free models depend on the same evidence: the gold-evidence gap is 0.063 for the zero-shot base, 0.097--0.119 under outcome-only GRPO and 0.067--0.071 for the plain mixed arms (\cref{tab:chainfree}, supplement). A positive \citegap therefore shows that the chain cites pixels the answer depends on, not that the chain is what makes it depend on them. And in a sparse scene a random box mostly covers background. On the items whose control lands on another object, the gap remains for the four models trained with CLEVRER chains (0.034--0.083, every interval above zero): their cited objects matter more than other objects. For the NExT-QA chain GRPO model, whose boxes are mostly off the named object (hit rate 0.15), it vanishes (0.007 \CI{-0.017}{0.033}), so that model's gap reflects objects against background (\cref{tab:clevrer_strata}, supplement). The \citegap speaks about the cited evidence only with a grounding check or an object-level control.

\subsection{NExT-QA: little depends on the cited box, more on the cited entity}
\label{sec:results_real}

\begin{figure*}[t]
\centering
\includegraphics[width=0.7\textwidth]{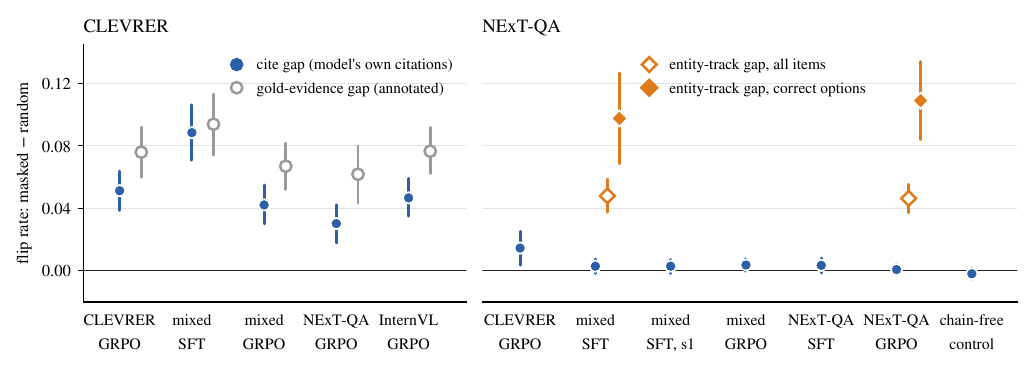}
\caption{\textbf{Cite gaps on CLEVRER and NExT-QA} (95\% question-clustered bootstrap intervals). Left: on CLEVRER the answers of all five tested chain models depend on their cited boxes, somewhat less than on the annotated evidence (hollow). Right: on NExT-QA the \citegap of the models trained on NExT-QA chains, of a second seed and of a chain-free control is at most 0.004; removing the cited entities from every frame (diamonds) gives a gap of 0.05 over all items and 0.10--0.11 on correct options.}
\label{fig:cite_gap}
\end{figure*}

\begin{table*}[t]
\centering\tabsize\setlength{\tabcolsep}{4pt}
\caption{\textbf{NExT-QA validation subset} (999 questions, 4,995 per-option items): accuracy and the self-citation test. Per-opt.\ is per-option accuracy (always-``no'' 0.80), Sel.\ picks the option judged ``yes'' and is comparable to the letters prompt, and the blind gap is in per-option accuracy (per-question for the letters row). Flip rates are over the items correct under the plain input; 95\% intervals in brackets. $^{\ddagger}$\,Control: this chain-free model is masked with the mixed chain SFT model's citations. The pooled row covers the five models trained on NExT-QA chains. Further accuracy columns are in the supplement (\cref{tab:nextqa}).}
\label{tab:real_tests}
\input{tables/tab_real_tests}
\end{table*}

\paragraph{Chain supervision does not change NExT-QA accuracy.} Models trained on CLEVRER chains alone keep their format on NExT-QA but not their judgment: they say ``yes'' to 38--41\% of options when 20\% are correct (\cref{tab:nextqa}, supplement). Pseudo-labelled NExT-QA chains repair this: the mixed and NExT-QA chain arms reach 0.84--0.85 per option and 0.64--0.65 selection accuracy (\cref{tab:real_tests}). The plain arm trained on the same questions with bare answers reaches the same numbers, and every trained arm is below the zero-shot base under a letters prompt (0.730); the per-option format accounts for the distance, since in it the untrained base drops to 0.595. The models depend on the video for a minority of their answers: with every frame black, 29\% of the chain models' correct answers flip, against 48\% on CLEVRER.

\paragraph{Little depends on the cited box.} For every model trained with NExT-QA chains the \citegap is 0.001--0.004, for supervised and RL-trained models alike and under a second SFT seed (\cref{tab:real_tests}, \cref{fig:cite_gap} right). Pooled over the five models, blacking out the cited boxes flips 2.3\% of correct answers against 2.1\% for the control: a gap of 0.003 \CI{0.001}{0.005} and a ratio of 1.1, where CLEVRER has 0.053 and 2.2. The gap is small, not zero. The flip rate itself bounds it under any control: at most 2.6\% of correct answers can be attributed to the cited boxes, against 9.7\% flipped on CLEVRER. Net accuracy under the cited mask moves by at most 0.004, and the selected answer changes for 6--10\% of questions under the mask and 5--11\% under the control (\cref{tab:symmetric}, supplement). The chain-free control gives $-0.002$. The CLEVRER chain GRPO model has a larger gap (0.015), from its rejections alone.

\paragraph{Checks on the small gap.} \emph{Video dependence.} Among the answers that flip when every frame is black, the cited boxes flip 4.5\% against 3.9\% for the control (0.006 \CI{0.001}{0.011}); on CLEVRER the same subset gives 12.4\% against 5.7\% (\cref{tab:vd}, supplement). \emph{Gold answer.} On correct options the pooled gap is 0.007 \CI{0.002}{0.012}, against 0.050 on CLEVRER. \emph{Mask size.} The cited boxes cover 3.5--4.1\% of pixel-frames on NExT-QA and 0.4--2.4\% on CLEVRER. \emph{Box placement.} The named entity is re-detected on the cited frame in 89--92\% of steps, and the cited box overlaps the detection with a hit rate of 0.60--0.62 at IoU 0.5, against 0.17--0.19 for a same-size random box (\cref{tab:grounding}, supplement); judged by hand, the box is on the entity in 52\% of 301 steps (\cref{tab:model_audit}, supplement). On the 29--35\% of cited items in which every box hits the detection, the \citegap is $-0.001$ to 0.011, with one interval of five above zero (\cref{tab:hit_split}, supplement). \emph{Control overlap.} NExT-QA boxes are large, so the random control covers 38--40\% of the cited area on average (6--10\% on CLEVRER). On the 17--21\% of items whose control leaves the cited area untouched, the gap is 0.000--0.006 with every interval including zero (\cref{tab:control}, supplement). \emph{Coverage.} The gap does not grow where the citations cover more of the entity's visible frames, except in one model of five (0.023 \CI{0.010}{0.037} at 30\% coverage or more; \cref{tab:dose}, supplement).

\begin{table}[t]
\centering\tabsize\setlength{\tabcolsep}{3.5pt}
\caption{\textbf{Entity-track test on NExT-QA.} Every detection of every entity the chain names, on all 16 frames, plus the cited boxes, is blacked out (Flip) against same-size boxes elsewhere on the same frames that avoid the entity boxes (Ctrl.); 95\% intervals in brackets.}
\label{tab:entity_track}
\input{tables/tab_entity_track}
\end{table}

\paragraph{More depends on the cited entity.} The named entities are detected on 14.0 of the 16 frames on average, so a citation of two frames removes little of them. The entity-track test (\cref{sec:faithfulness}) removes every detection on all 16 frames plus the cited boxes, 8.7\% of the pixel-frames, against the same boxes placed off the entity (\cref{tab:entity_track}). Pooled over the two models it was run on, it flips 10.2\% of correct answers against 5.5\% for the control, a gap of 0.047 and a ratio of 1.9; on correct options the gap is 0.103, and the selected answer changes for 35--36\% of questions against 19--22\%. Both gaps exceed the bound on the box-level effect, so the difference between the two tests is not an artefact of their controls. Nor is it an artefact of entities copied from the question, or of a control that only removes background: the gap is as large on the items none of whose named entities shares a word with the question or the options (0.046--0.049; \cref{tab:text_split}, supplement), and it remains where the control mostly lands on another detected entity (0.037 and 0.044, both intervals above zero; \cref{tab:entity_control}, supplement). Relative to the blind flip rate, the pooled gaps are 0.01 for the cited boxes on NExT-QA and 0.11 on CLEVRER, 0.17 for the entity track, and 1.0 for the CLEVRER object track, which removes annotated decisive objects. We did not run the entity-track test on a chain-free model, so it shows that the answer depends on the entities the chain names more than on other entities, not that the chain is the reason.

%% file: tables/tab_cite_clevrer.tex
\begin{tabular}{@{}l ccc c cc c c@{}}
\toprule
Model & Acc. & Blind gap & Hit@0.5 & $n$ & Flip (cited) & Flip (random) & Cite gap & Gold-evidence gap \\
\midrule
CLEVRER chain GRPO & 0.871 & 0.37 & 0.78 & 2527 & 0.088 & 0.036 & 0.051\,{\scriptsize[0.039,0.064]} & 0.076\,{\scriptsize[0.060,0.092]} \\
Mixed chain SFT & 0.739 & 0.27 & 0.78 & 2512 & 0.165 & 0.077 & 0.089\,{\scriptsize[0.071,0.106]} & 0.094\,{\scriptsize[0.074,0.113]} \\
Mixed chain GRPO & 0.771 & 0.29 & 0.77 & 2520 & 0.078 & 0.036 & 0.042\,{\scriptsize[0.030,0.055]} & 0.067\,{\scriptsize[0.052,0.082]} \\
NExT-QA chain GRPO & 0.535 & 0.08 & 0.15 & 2533 & 0.069 & 0.038 & 0.030\,{\scriptsize[0.018,0.042]} & 0.062\,{\scriptsize[0.043,0.080]} \\
InternVL3-2B chain GRPO & 0.803 & 0.33 & 0.76 & 2532 & 0.084 & 0.037 & 0.047\,{\scriptsize[0.035,0.059]} & 0.077\,{\scriptsize[0.062,0.092]} \\
\bottomrule
\end{tabular}

%% file: tables/tab_real_tests.tex
\begin{tabular}{@{}l ccc cc cc@{}}
\toprule
Model & Per-opt. & Sel. & Blind gap & Flip (cited) & Flip (random) & Cite gap & Cite gap, correct opt. \\
\midrule
Qwen2.5-VL-3B, letters prompt & -- & 0.730 & 0.29 & -- & -- & -- & -- \\
Qwen2.5-VL-3B & 0.860 & 0.595 & 0.06 & -- & -- & -- & -- \\
Mixed plain SFT (chain-free)$^{\ddagger}$ & 0.826 & 0.637 & 0.12 & 0.014 & 0.016 & $-$0.002\,{\scriptsize[$-$0.005,0.001]} & 0.002\,{\scriptsize[$-$0.005,0.010]} \\
Mixed plain SFT, seed 1 & 0.849 & 0.661 & 0.13 & -- & -- & -- & -- \\
\midrule
CLEVRER chain GRPO & 0.682 & 0.470 & 0.12 & 0.085 & 0.070 & 0.015\,{\scriptsize[0.003,0.025]} & 0.000\,{\scriptsize[$-$0.019,0.019]} \\
Mixed chain SFT & 0.845 & 0.637 & 0.22 & 0.028 & 0.025 & 0.003\,{\scriptsize[$-$0.002,0.007]} & 0.012\,{\scriptsize[$-$0.001,0.026]} \\
Mixed chain SFT, seed 1 & 0.846 & 0.642 & 0.16 & 0.023 & 0.020 & 0.003\,{\scriptsize[$-$0.002,0.007]} & 0.003\,{\scriptsize[$-$0.010,0.015]} \\
Mixed chain GRPO & 0.837 & 0.640 & 0.25 & 0.022 & 0.018 & 0.004\,{\scriptsize[0.000,0.008]} & 0.010\,{\scriptsize[0.001,0.018]} \\
NExT-QA chain SFT & 0.837 & 0.641 & 0.19 & 0.029 & 0.025 & 0.003\,{\scriptsize[$-$0.001,0.008]} & 0.006\,{\scriptsize[$-$0.005,0.017]} \\
NExT-QA chain GRPO & 0.852 & 0.653 & 0.15 & 0.016 & 0.015 & 0.001\,{\scriptsize[$-$0.003,0.004]} & 0.005\,{\scriptsize[$-$0.004,0.014]} \\
\midrule
Pooled, last five rows &  &  &  & 0.023 & 0.021 & 0.003\,{\scriptsize[0.001,0.005]} & 0.007\,{\scriptsize[0.002,0.012]} \\
\bottomrule
\end{tabular}

%% file: tables/tab_entity_track.tex
\begin{tabular}{@{}ll cc c@{}}
\toprule
Model & Items & Flip & Ctrl. & Gap \\
\midrule
Mixed chain SFT & All & 0.110 & 0.062 & 0.048\,{\scriptsize[0.038,0.059]} \\
 & Correct opt. & 0.194 & 0.096 & 0.098\,{\scriptsize[0.069,0.127]} \\
 & Wrong opt. & 0.090 & 0.054 & 0.037\,{\scriptsize[0.026,0.048]} \\
\midrule
NExT-QA chain GRPO & All & 0.095 & 0.049 & 0.046\,{\scriptsize[0.037,0.055]} \\
 & Correct opt. & 0.167 & 0.058 & 0.109\,{\scriptsize[0.084,0.134]} \\
 & Wrong opt. & 0.078 & 0.046 & 0.031\,{\scriptsize[0.023,0.040]} \\
\midrule
Pooled & All & 0.102 & 0.055 & 0.047\,{\scriptsize[0.039,0.056]} \\
 & Correct opt. & 0.180 & 0.076 & 0.103\,{\scriptsize[0.081,0.127]} \\
 & Wrong opt. & 0.084 & 0.050 & 0.034\,{\scriptsize[0.026,0.042]} \\
\bottomrule
\end{tabular}

%% file: sec/6_discussion.tex
\section{Discussion}
\label{sec:discussion}

\paragraph{What the tests show.} The \citegap and the entity-track gap measure dependence at the input: they compare losing the cited pixels with losing an equal area of other pixels. They do not show that the answer is computed \emph{through} the chain: on CLEVRER chain-free models depend on the annotated evidence as much as chain models do, and we did not run the entity-track test on a chain-free model or perturb the chain text. They are not sufficiency tests, since we never keep only the cited regions. The control is a random position; object-level and entity-level controls were obtained only by replaying it, not by a separate run. And a black box is out of distribution: the control alone flips 5--6\% of correct answers at 8.7\% masked area. A gap licenses a graded statement, that the answer depends on what the chain points at more than on other pixels by so much, and the intervention's own flip rate bounds it. A small gap has two readings, which our tests do not separate: the model does not use the cited box, or it uses it and finds the same content elsewhere. Under the cited mask the NExT-QA chains re-cite the blacked-out box in only 14--17\% of steps and move to another box on the same frame in 67--74\% (\cref{tab:recite}, supplement), which fits the second reading. The citations are also unstable in the sense of~\cite{subramaniakuppusamy2026feature}: under the random mask the answer is unchanged for 96--98\% of items, and on those the same box is cited again in only 52--55\% of steps (76--84\% on CLEVRER).

\paragraph{Why so little depends on the cited box on NExT-QA.} CLEVRER rules out the simplest account, that a 3B model cannot cite evidence its answer depends on. For NExT-QA the supervision predicts the outcome. The pseudo-labels cite two frames, and the entities the trained models name are detected on fourteen of sixteen; the same steps are attached to the correct option and to a wrong one, and in our audit 92 of 100 training chains only locate and describe an entity, written by a teacher that was shown the answer. Nothing ties the answer to the box: the bridge loss (\cref{eq:bridge}) corrupts a step's intermediate answer, not its evidence. The two benchmarks thus differ in three ways at once, redundancy of the evidence, citation format and supervision content, and our data separate them only in part: the entity-track test shows that redundancy accounts for a gap of 0.05, and nothing here isolates the other two; within NExT-QA the \citegap does not rise with the share of the entity's frames that the citations cover. A crossed design (pseudo-labelled chains on CLEVRER, track-level spans on NExT-QA) would.

\paragraph{What the case is for.} The value of the NExT-QA result is not that it surprises. It is that the standard evaluation does not reveal it: on CLEVRER and on NExT-QA alike the chains parse, the boxes pass a grounding check, and the models use the video, while the dependence of the answer on the cited box differs by an order of magnitude. Work that reports grounded chains should report a \citegap and, where evidence is redundant, an entity-track gap beside it.

\paragraph{Limitations.} All training is LoRA on 2--3B models with at most 12,400 chain items and 1,000--1,500 RL steps. The NExT-QA results come from one benchmark and one base (InternVL3-2B was run on CLEVRER only); the entity-track test covers two models; no released grounded video model was tested. Only the mixed chain SFT arm has two seeds on NExT-QA, and intervals cover items, not training variance. The bridge loss is not ablated. The pseudo-labelled chains are noisy (box on the entity in 70\% of audited steps, one annotator) and observational. The grounding check and the entity-track test use the same detector family as the training labels, and the latter removes only what the detector finds of what the chain names.

%% file: sec/7_conclusion.tex
\section{Conclusion}
A grounded reasoning chain makes two promises: that its citations can be checked, and that its answer rests on them. We proposed a counterfactual self-citation test for a consequence of the second promise, with an entity-track extension for redundant evidence, and applied both to small video--language models trained to reason in grounded chains. On CLEVRER, masking the cited boxes flips about twice as many correct answers as masking equal-area regions elsewhere, for every chain model on two bases. On NExT-QA, where the chains were supervised to cite two frames of a persistent entity, masking the cited boxes flips 2.3\% of correct answers against 2.1\%, and masking the named entity on every frame costs ten points more than a size-matched control on the options the model correctly affirms. Both sets of boxes pass a grounding check. Checkable, relevant citations can be ones the answer does not depend on, and both tests, which need no annotation, should accompany any claim that a video model's grounded reasoning explains its answers. Code, data and weights will be released at \href{https://github.com/JugalGajjar/CausalSight}{\texttt{github.com/JugalGajjar/CausalSight}}.

%% file: sec/X_suppl.tex
\clearpage
\maketitlesupplementary
\appendix
\crefalias{section}{appendix}\crefalias{subsection}{subappendix}%
\setcounter{table}{0}\renewcommand{\thetable}{S\arabic{table}}
\setcounter{figure}{0}\renewcommand{\thefigure}{S\arabic{figure}}

\noindent This supplement gives the full chains behind \cref{fig:teaser} (\cref{sec:supp_examples}); CLEVRER accuracy for every model (\cref{sec:supp_clevrer_acc}); pooled estimates and bounds for both benchmarks (\cref{sec:supp_pooled}); the per-type, per-model and per-answer numbers behind the self-citation and entity-track tests and their diagnostics (\cref{sec:supp_bytype,sec:supp_nextqa,sec:supp_cite_nq,sec:supp_cite_track,sec:supp_grounding,sec:supp_gold,sec:supp_diag}); the manual audits of the pseudo-labels and of the models' own citations (\cref{sec:supp_audit}); the remaining results of the mixed-data experiment and the chain-free models (\cref{sec:supp_mixed}); two side results, on RL collapse (\cref{sec:supp_collapse}) and on process rewards (\cref{sec:supp_stage2}); the second base model (\cref{sec:supp_internvl}); CausalVQA (\cref{sec:supp_causalvqa}); and training, compute and prompt details (\cref{sec:supp_training,sec:supp_prompt}).

\section{Example chains}
\label{sec:supp_examples}
\begin{figure*}[t]
\centering
\includegraphics[width=\textwidth]{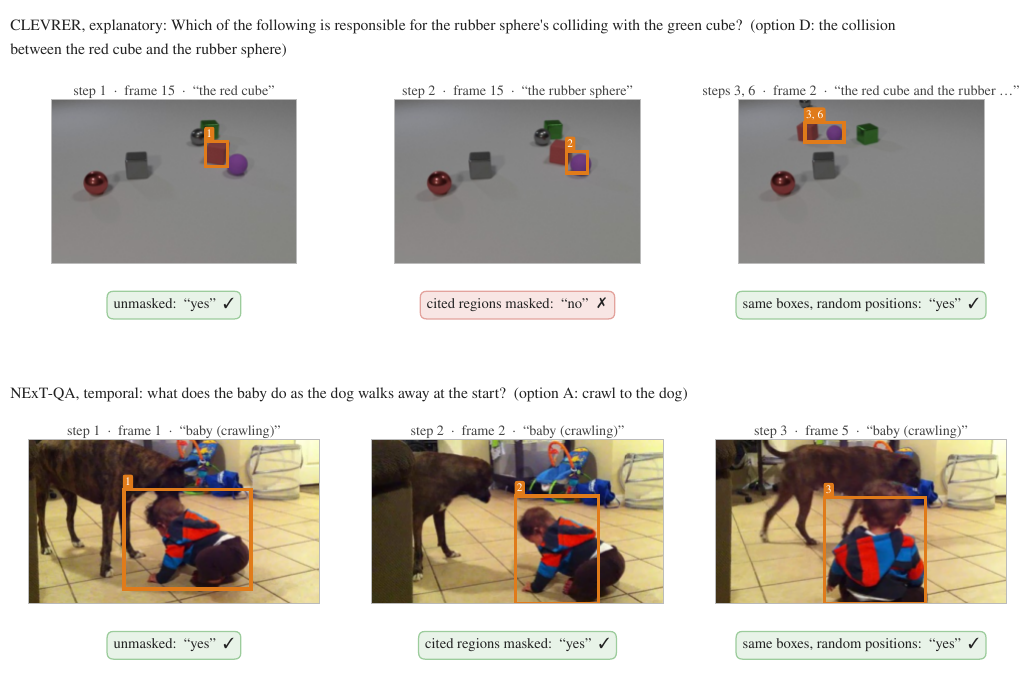}
\caption{\textbf{Two chains from the mixed chain SFT model.} Top: a CLEVRER explanatory item; masking the cited regions flips the answer, masking boxes of the same size elsewhere does not. Bottom: a NExT-QA temporal item; the cited boxes sit on the named entity in every step and the answer survives masking them. Only steps with distinct cited regions are shown.}
\label{fig:examples}
\end{figure*}

\cref{fig:examples} shows the full chains behind \cref{fig:teaser}: each step with a distinct cited region, the frame it points to, and the model's answer under the three conditions of the self-citation test. The CLEVRER chain resolves the two objects named in the option and then cites the frame of their collision; masking those regions flips the answer and masking equal-sized regions elsewhere does not. The NExT-QA chain cites the baby on three frames, correctly, and the answer survives masking all three boxes.

\section{Pooled estimates and bounds}
\label{sec:supp_pooled}
\cref{tab:pooled} puts the two benchmarks side by side with one statistic. Items of several models are pooled and a question's items from every model form one bootstrap cluster. Flip is the share of correct answers that become wrong under the intervention, with its own interval: since a gap cannot exceed the flip rate under any control, its upper end bounds what the intervened pixels can account for. On CLEVRER the cited boxes flip 9.7\% of correct answers, 2.2 times the control. On NExT-QA they flip 2.3\% \CI{0.021}{0.026}, 1.1 times the control; the pooled gap of 0.003 is small and its interval excludes zero. The entity-track test on NExT-QA has the size of the box-level test on CLEVRER, and both are far below the object-track test on CLEVRER, which removes annotated decisive objects.

\begin{table*}[h]
\centering\tabsize
\caption{\textbf{Pooled tests on both benchmarks.} $n$ = pooled items correct under the plain input; Flip and Ctrl.\ are the shares that become wrong under the intervention and under its size-matched control; Ratio is Flip over Ctrl. CLEVRER rows for correct options count descriptive questions with a ``yes'' answer and leave out the open-ended ones.}
\label{tab:pooled}
\input{tables/tab_pooled}
\end{table*}

\paragraph{Video-dependent answers.} A model that answers from the question alone cannot flip under any mask. \cref{tab:vd} restricts every test to the \emph{video-dependent} answers, those that are correct on the unmodified input and wrong when every frame is black, and adds the pooled gap over all items as a share of the blind flip rate. With every frame black, 29\% of the correct NExT-QA answers flip and 48\% of the CLEVRER ones. Among those, the cited boxes flip 4.5\% against 3.9\% on NExT-QA and 12.4\% against 5.7\% on CLEVRER. As a share of the blind flip rate, the cited boxes account for 1\% on NExT-QA and 11\% on CLEVRER, the entity track for 17\%, and the object track on CLEVRER for all of it. The small NExT-QA \citegap is therefore not a by-product of answers that ignore the video.

\begin{table*}[h]
\centering\tabsize
\caption{\textbf{The tests on video-dependent answers}, pooled. Blind flip: share of correct answers that flip when every frame is black. $n$, Flip, Ctrl.\ and Gap are over the answers that do. The last column is the gap over all correct answers (\cref{tab:pooled}) divided by the blind flip rate.}
\label{tab:vd}
\input{tables/tab_vd}
\end{table*}

\section{Per-type cite gaps}
\label{sec:supp_bytype}
\cref{tab:cite_bytype} breaks the \citegap down by question type. On CLEVRER the pattern is not uniform across models. The mixed chain SFT has its largest gaps on explanatory and predictive questions, whose evidence is one collision at one place, and a descriptive gap that includes zero; the CLEVRER chain GRPO and InternVL models have their largest gap on predictive questions; and the NExT-QA chain GRPO model has its one large gap on descriptive questions (0.129). On NExT-QA no per-type gap of a model trained on NExT-QA chains exceeds 0.008, and one of the fifteen intervals excludes zero.

\begin{table*}[h]
\centering\tabsize
\caption{\textbf{Cite gap by question type}, with 95\% intervals. Upper block: CLEVRER (descriptive, explanatory, predictive, counterfactual). Lower block: NExT-QA (causal, temporal, descriptive).}
\label{tab:cite_bytype}
\input{tables/tab_cite_bytype}
\end{table*}

\section{NExT-QA accuracy: all arms}
\label{sec:supp_nextqa}
\cref{tab:nextqa} gives every NExT-QA arm with per-question accuracy, the ``yes'' rate and blind selection accuracy, which \cref{tab:real_tests} omits. The reference models appear in both formats because the per-option format itself costs accuracy: the untrained base falls from 0.730 with a letters prompt to 0.595 selection accuracy.

\begin{table*}[h]
\centering\tabsize
\caption{\textbf{NExT-QA validation subset} (999 questions, 4,995 per-option items; chain-style models answer one yes/no item per option). Acc.\ (q): all five options judged correctly; selection picks the option judged ``yes'' (comparable to the letters prompt); blind gap is in per-option accuracy (per-question for the letters rows). Seed-1 rows are a second SFT run of the same recipe. The two letters-prompt rows have the same blind accuracy by coincidence (437 of 999 questions each, on different outputs).}
\label{tab:nextqa}
\input{tables/tab_nextqa}
\end{table*}

\section{Self-citation test on NExT-QA: full numbers}
\label{sec:supp_cite_nq}
\cref{tab:cite_nextqa} lists the numbers plotted in \cref{fig:cite_gap} (right). The flip rates themselves are informative: blacking out the cited regions flips 1.6--2.9\% of correct answers, blacking out the same area elsewhere 1.5--2.5\%, so the cited regions are close to interchangeable with any other region of the same size.

\begin{table*}[h]
\centering\tabsize
\caption{\textbf{Self-citation test on NExT-QA} (4,995 per-option items). Same statistic as \cref{tab:cite_clevrer}; $n$ counts the items whose chain cites a localisable region, and the flip rates are over the correct ones among them. The last row masks a chain-free model, which writes no chains, with the mixed chain SFT model's citations: a check that the masking operation is neutral. The intervals of the models trained on NExT-QA chains include or touch zero; pooled over them the interval is above zero (\cref{tab:pooled}).}
\label{tab:cite_nextqa}
\input{tables/tab_cite_nextqa}
\end{table*}

\section{Entity-track test on NExT-QA}
\label{sec:supp_cite_track}
\cref{tab:cite_track} gives the entity-track test (\cref{sec:faithfulness}) for the two models it was run on, the mixed chain SFT and the NExT-QA chain GRPO model. The two chain models name 2,746 distinct video--phrase pairs over the 485 videos of the subset; Grounding DINO (tiny, threshold 0.35) detects a phrase on 14.0 of the 16 sampled frames on average, on eight or more frames for 94.0\% of phrases, and never for 1.5\%. Items whose named entities are never detected are dropped (133 of the 4,220 mixed chain SFT items correct under the plain input; 114 of 4,254 for the NExT-QA chain GRPO model). The intervention blacks out 8.7\% of the pixel-frames (median 8.3\%) against 3.9\% for the cited boxes alone; the control blacks out the same area at the same times, placed to avoid the entity boxes. Split by detector coverage, the items whose entities were detected on 14 or more frames on average have gaps of 0.053 (mixed chain SFT, 3,088 items) and 0.050 (NExT-QA chain GRPO, 3,287), and the remaining 999 and 853 items 0.032 and 0.033, so the gap with full detector coverage would be close to the one reported, not several times larger.

\begin{table*}[h]
\centering\tabsize\setlength{\tabcolsep}{4.2pt}
\caption{\textbf{Entity-track test on NExT-QA.} Every detection of every entity the chain names, on all 16 frames, plus the cited boxes, is blacked out (Flip, tracks) against same-size boxes elsewhere on the same frames (Flip, random); $n$ counts the items correct under the plain input whose entities were detected. The cite gap and blind gap of the same model are repeated for comparison; 95\% intervals in brackets.}
\label{tab:cite_track}
\input{tables/tab_cite_track_nextqa}
\end{table*}

\section{Citation grounding on NExT-QA}
\label{sec:supp_grounding}
\cref{tab:grounding} gives the detector check of \cref{sec:faithfulness} for every chain model. The CLEVRER chain GRPO model writes few ``Where is'' steps.

\begin{table*}[h]
\centering\tabsize
\caption{\textbf{Do the NExT-QA citations agree with a detector?} Each ``Where is $\langle$entity$\rangle$?'' entity is re-detected with Grounding DINO on the cited frame; Det.\ is the share of steps with a detection; IoU and Hit@0.5 compare the cited box (cited / same-size random box) with the top detection.}
\label{tab:grounding}
\input{tables/tab_grounding}
\end{table*}

\section{Correct and wrong options}
\label{sec:supp_gold}
With one yes/no item per option, four of five NExT-QA items are wrong options, and a correct rejection rarely flips under a box mask. \cref{tab:gold_split} therefore repeats both tests separately for the items whose option is correct and wrong. The small \citegap on NExT-QA is not a dilution effect: on correct options the \citegap of the models trained on NExT-QA chains is 0.003--0.012, and one interval of five excludes zero. The entity-track gap, in contrast, is concentrated on correct options (0.098 and 0.109 against 0.037 and 0.031): removing the entities a chain names mainly withdraws the model's support for affirming an option. On the same correct affirmations, blacking out every frame flips 20.9\% (mixed chain SFT) and 34.3\% (NExT-QA chain GRPO). On CLEVRER the \citegap is positive on correct and on wrong options for every model, so there the cited evidence moves the answer in both directions; it is larger on wrong options for the mixed chain GRPO model (0.074 against 0.029) and larger on correct options for InternVL (0.055 against 0.034).

\begin{table*}[h]
\centering\tabsize\setlength{\tabcolsep}{4.5pt}
\caption{\textbf{Both tests by gold answer.} $n$ = items of that kind that are correct under the plain input; Flip and Ctrl.\ are the shares of them that become wrong under the intervention and under its size-matched control; 95\% intervals in brackets. On CLEVRER, descriptive questions with a yes/no answer are counted by that answer and the other descriptive questions are left out.}
\label{tab:gold_split}
\input{tables/tab_gold_split}
\end{table*}

\section{Citation hits, masked area and re-citation}
\label{sec:supp_diag}
\paragraph{Citation hits.} \cref{tab:hit_split} splits the NExT-QA items whose chain cites a region by whether every ``Where is $\langle$entity$\rangle$?'' box of the plain chain overlaps the detector's box for that entity on the cited frame at IoU 0.5 or more (a step without a detection counts as a miss). Where every box hits, the \citegap is $-0.001$ to 0.011, and one interval of five excludes zero; the entity-track gap is the same in both groups. The small \citegap is therefore not produced by boxes in the wrong place.

\begin{table*}[h]
\centering\tabsize
\caption{\textbf{Gaps by citation hit on NExT-QA.} Left block: items in which every cited box overlaps the detector's box for the named entity at IoU $\ge0.5$; right block: the other items. 95\% intervals in brackets.}
\label{tab:hit_split}
\input{tables/tab_hit_split}
\end{table*}

\paragraph{Citation correctness and an object-level control on CLEVRER.} In a sparse scene a uniformly random box mostly covers background, so a positive \citegap could mean ``objects against floor''. \cref{tab:clevrer_strata} replays the control draws and separates the items whose control box covers at least 30\% of an annotated object from those whose control lands on background only. The control flips more answers when it removes an object (4.8--10.5\% against 1.4--3.5\%), and for the four models trained with CLEVRER chains the \citegap remains positive against that stricter control (0.034--0.083, every interval above zero). For the NExT-QA chain GRPO model, whose boxes are mostly off the object they name, it does not (0.007 \CI{-0.017}{0.033}): that model's gap of 0.030 over all items is an object-against-background effect. The table also splits the items by whether every cited box overlaps the object it names at IoU 0.5 or more; the gap is of similar size in both groups, so on CLEVRER it is not confined to perfectly placed citations.

\begin{table*}[h]
\centering\tabsize\setlength{\tabcolsep}{3pt}
\caption{\textbf{CLEVRER \citegap by citation correctness and by where the control lands.} Left: items in which every cited box is on the object it names (IoU $\ge0.5$) against the rest, among chains that name at least one fully specified object. Right: items whose random control covers at least 30\% of an annotated object against items whose control covers background only, with the control's flip rate. $n$ = correct items in the group.}
\label{tab:clevrer_strata}
\input{tables/tab_clevrer_strata}
\end{table*}

\paragraph{Reverse flips, net accuracy and question-level flips.} The gaps count correct answers that become wrong. \cref{tab:symmetric} adds the other direction and two coarser views. On NExT-QA the cited mask also turns wrong answers right, slightly more often than the control, so net per-option accuracy moves by at most 0.004 under either; on CLEVRER it falls by 0.04--0.07 under the cited mask and 0.01--0.02 under the control for the models trained with CLEVRER chains. At the level of questions, the selected answer of a question whose five options were all judged correctly changes for 6--10\% of questions under the cited mask and 5--11\% under the control, and for 35--36\% against 19--22\% under the entity-track mask.

\begin{table*}[h]
\centering\tabsize
\caption{\textbf{Net accuracy, reverse flips and question-level flips.} Net acc.: per-option accuracy under the mask (or its control) minus accuracy on the unmodified input. W$\to$C: share of wrong answers that become correct. Question flip: share of questions with a correct selection on the unmodified input whose selection is no longer correct. The entity-track runs cover only items that are correct on the unmodified input, so their net change equals minus the flip rate.}
\label{tab:symmetric}
\input{tables/tab_symmetric}
\end{table*}

\paragraph{Does the gap grow where less of the entity survives the mask?} If the \citegap on NExT-QA were small only because the entity remains visible on other frames, it should grow where the citations cover more of the frames on which the entity is detected. \cref{tab:dose} bins the items by that share. For four of five models the gap is flat; for the mixed chain SFT it reaches 0.023 \CI{0.010}{0.037} where the citations cover 30\% or more. The entity-track gap, in contrast, is present for entities detected on nine or more frames and absent or uncertain for the few detected on eight or fewer. Redundancy is therefore shown by the entity-track test, not by a within-benchmark trend of the \citegap.

\begin{table*}[h]
\centering\tabsize\setlength{\tabcolsep}{5pt}
\caption{\textbf{Dose--response on NExT-QA.} Upper block: \citegap by the share of the frames showing the named entities that the citations cover. Lower block: entity-track gap by the number of frames (of 16) on which the named entities are detected. 95\% intervals in brackets.}
\label{tab:dose}
\input{tables/tab_dose}
\end{table*}

\paragraph{Control overlap.} The self-citation control places each box uniformly at random on the same frames and may overlap the cited box. \cref{tab:control} replays the seeded control draws and reports the mean share of the cited area that the control also blacks out, the share of items whose cited area the control leaves untouched, and the flip rate and \citegap on those items. On CLEVRER the control covers 6--10\% of the cited area. On NExT-QA the boxes are large and it covers 38--40\%, so on the full item set the control is a weaker contrast there. On the items with an untouched cited area the NExT-QA gap is 0.000--0.006 with every interval including zero, and the cited-box flip rate is 1.5--2.6\%; on CLEVRER the same subset has gaps of 0.036--0.093. Overlap does not produce the small NExT-QA gap. The table also gives the number of sampled frames a cited region is applied to. On NExT-QA it is 2.0 for every model trained with NExT-QA chains and 1.3 for the CLEVRER chain GRPO model; on CLEVRER it is 1.2--1.5 for the models trained with CLEVRER chains and 2.0 for the NExT-QA chain GRPO model. Span length follows the training chains, not the benchmark.

\begin{table*}[h]
\centering\tabsize
\caption{\textbf{Control overlap.} Frames: sampled frames a cited region is applied to. Covered: mean share of the cited area also blacked out by the random control. Untouched: share of items whose cited area the control does not touch; $n$, flip rate and \citegap are for the correct ones among them.}
\label{tab:control}
\input{tables/tab_control}
\end{table*}

\paragraph{Entities from the question text.} A chain may name the question's own subject, and removing that would hurt any model. \cref{tab:text_split} compares the items in which every named entity has all its content words in the question or option text (39--45\% of items) with the items in which no named entity shares a content word with that text (35--38\%). The entity-track gap is as large in the second group as in the first (0.046 and 0.049 against 0.052 and 0.038; on correct options 0.115 and 0.120 against 0.081 and 0.086), so it is not carried by entities copied from the question.

\begin{table*}[h]
\centering\scriptsize\setlength{\tabcolsep}{3pt}
\caption{\textbf{Gaps by whether the named entities appear in the question or option text.} Left block: every named entity appears in the text; right block: no named entity shares a content word with it. E.-t.: entity-track gap. 95\% intervals in brackets.}
\label{tab:text_split}
\input{tables/tab_text_split}
\end{table*}

\paragraph{An entity-level control for the entity-track test.} The entity-track control avoids the named entity, so it could be comparing the video's main subject with background, the confound that \cref{tab:clevrer_strata} exposes for one model on CLEVRER. \cref{tab:entity_control} replays the seeded control draws and scores each item by the share of its control boxes that cover at least 30\% of \emph{another} detected entity on the same frame: a phrase named by a chain on the same video but not by this item's chain, not overlapping the item's own entity boxes, and smaller than 40\% of the frame. For about a fifth of the items the control lands on another entity on half or more of its boxes. There the control flips more answers for one model (9.5\% against 5.3\%) and slightly more for the other, and the entity-track gap remains: 0.037 \CI{0.011}{0.064} and 0.044 \CI{0.024}{0.067} over all items, 0.070 and 0.111 on correct options, every interval above zero. The gap is therefore not an entity-against-background effect: the entities a chain names matter more than other entities in the same video. It remains untested whether a chain-free model shows the same.

\begin{table*}[h]
\centering\tabsize
\caption{\textbf{Entity-track gap by where the control lands.} Items are grouped by the share of their replayed control boxes that cover at least 30\% of another detected entity: half or more (mostly on another entity), a tenth or less (background), or in between. $n$ = correct items in the group; 95\% intervals in brackets.}
\label{tab:entity_control}
\input{tables/tab_entity_control}
\end{table*}

\paragraph{Masked area and re-citation.} \cref{tab:recite} gives the share of pixel-frames blacked out by the self-citation intervention, and what the chain written \emph{under} a mask cites, relative to the boxes cited on the unmodified input: the same box (IoU $\ge0.5$ on a frame the old box was applied to), another place on such a frame, or a frame the old citations did not touch. The random-mask run is the baseline, since there the originally cited region is still visible. On NExT-QA the models trained on NExT-QA chains re-cite the blacked-out box in 14--17\% of steps against 52--55\% under the random mask, and name the same entities in 68--74\% of chains against 73--77\%: they keep the entity and move the box. On CLEVRER the chain models re-cite the blacked-out box more often (44--58\% against 75--83\%) and shift to other frames more (12--27\% against 6--12\%). In both domains where the model points depends on the pixels there; only on CLEVRER does the answer. The random-mask rates are a stability measurement, and stability should be read only where the prediction is preserved~\cite{subramaniakuppusamy2026feature}. Restricted to the items whose answer is unchanged under the random mask (96--98\% of items on NExT-QA, 90--96\% on CLEVRER), the same box is cited again in 52--55\% of steps on NExT-QA and 76--84\% on CLEVRER, the same as over all items: the citations move although the answer does not.

\begin{table*}[h]
\centering\tabsize\setlength{\tabcolsep}{4pt}
\caption{\textbf{Masked area and what the masked run cites.} Masked: mean share of pixel-frames blacked out by the cited boxes, in per cent. Columns 3--6: shares of the steps written under the cited-box mask that cite the blacked-out box again, another place on a masked frame, or another frame, and the share of chains that name the same entities as on the unmodified input; columns 7--10: the same under the random mask.}
\label{tab:recite}
\input{tables/tab_recite}
\end{table*}

\section{Manual audits of the pseudo-labels and of the models' citations}
\label{sec:supp_audit}
One author judged 100 pseudo-labelled NExT-QA training chains, sampled uniformly at random with one chain per question (299 steps; \cref{tab:audit}). Each step was shown as the cited frame (the sampled frame nearest the centre of the span) with the detector box drawn, next to the teacher's entity and observation, and judged yes, no or unsure on three questions; each chain was judged on two. The entity the teacher names is visible on the cited frame in 82\% of steps (246 of 299) and the detector box is on it in 70\% (209; 206 of the 246 steps where it is judged visible, 84\%); 51 chains have every step grounded and 14 have none. The recurring failures are a box on the wrong one of two people, an entity that is not on the cited frame, a stock phrase for an entity that is not in the video (``the man in the blue sweater''), an entity that is a predicate rather than an object (``walking on the path''), and boxes that cover the whole frame. Content is better than localisation: 95\% of steps are relevant to the question and 89 chains would suffice to answer it if their observations were taken as true. But the observations restate the answer the teacher was shown rather than derive it: only 8 chains contain a step that is more than locating an entity and describing it. The audit therefore measures two things the detector-agreement check of \cref{tab:grounding} cannot: the label noise the models were trained on, and the absence of causal structure in the supervision.

\begin{table}[h]
\centering\tabsize
\caption{\textbf{Manual audit of 100 pseudo-labelled NExT-QA chains} (299 steps, one annotator). Shares of yes / no / unsure.}
\label{tab:audit}
\input{tables/tab_audit}
\end{table}

\paragraph{The models' own citations.} The same author judged, on the same five judgements, 100 chains that two trained models wrote on NExT-QA validation items: 50 by the mixed chain SFT model and 50 by the NExT-QA chain GRPO model, on the same 50 questions (one item per question, drawn uniformly among the items whose chain cites a region; 301 steps; 86 of the 100 items are wrong options). \cref{tab:model_audit} sets them beside the pseudo-labels. The entity a step names is visible on the cited frame about as often as in the supervision (80\% of steps, 242 of 301, against 82\%), but the box is on it less often: 52\% of steps (158), and 65\% of the 242 steps where the entity is visible, against 70\% and 84\%. The two models do not differ (52\% and 53\%). Of the 100 chains, 35 have every step on its entity and 28 have none. Three patterns recur: boxes that cover at least 80\% of the frame (42 steps, 29 of them judged unsure), a box beside the entity or on the wrong one of several people, and an entity that is not on the cited frame. The audit agrees with the detector check of \cref{tab:grounding} without relying on the detector: there, 90\% and 91\% of the two models' steps have a detection and 60\% and 62\% of those are hit at IoU 0.5, that is 54\% and 56\% of all steps.

The content of the chains mirrors the supervision. Every step is relevant to the question, 32 chains repeat one observation in all of their steps, and 5 chains contain a step that is more than locating and describing an entity (8 of 100 in the pseudo-labels). Taken as true, the observations would settle the item in 51 chains (33 no, 16 unsure). This is not comparable with the 89 of the first audit, whose chains were written for the correct option by a teacher that had been shown it; here four of five items ask about a wrong option, which a chain can only reject by saying what happened instead.

Each item is a yes/no verdict on one option, so a second sheet stated the model's verdict next to each chain and asked whether the steps, taken as true, support it (last row of \cref{tab:model_audit}). They support it in 63 of the 100 chains, lead to the opposite verdict in 3, and do not bear on the option in 34 (31, 2 and 17 for the mixed chain SFT model; 32, 1 and 17 for the NExT-QA chain GRPO model). In this sample the share is the same whether the verdict is right (50 of 79) or wrong (13 of 21): the chains rarely contradict the answer, and a chain that supports the answer does not indicate that the answer is correct.

\begin{table}[h]
\centering\tabsize
\caption{\textbf{Manual audit of the models' own citations.} Share of ``yes'' (number judged). Labels: the pseudo-labels of \cref{tab:audit}. SFT and GRPO: 50 chains each from the mixed chain SFT and the NExT-QA chain GRPO model, on the same 50 NExT-QA validation questions. The last row was judged on a separate sheet that stated the model's verdict. One annotator.}
\label{tab:model_audit}
\input{tables/tab_model_audit}
\end{table}

\section{CLEVRER accuracy}
\label{sec:supp_clevrer_acc}
\cref{tab:clevrer_main} gives the CLEVRER accuracy, grounding and video-dependence numbers summarised in \cref{sec:results_clevrer}, for the zero-shot base, outcome-only GRPO, three released checkpoints and the chain models. Outcome-only GRPO was trained on two mixes of 1,500 CLEVRER questions: \emph{balanced} (375 per question type) and \emph{natural} (CLEVRER's own mix, 72\% descriptive).

\begin{table*}[h]
\centering\tabsize\setlength{\tabcolsep}{3.4pt}
\caption{\textbf{CLEVRER validation, matched preprocessing.} Descriptive accuracy is per question; the other types are per option (yes/no per choice); All is over the 2,537 per-option items and is given for the models that answer one item per option. Box IoU is the mean IoU between a step's cited box and the object it names. Blind gap is a difference in accuracy and track gap (the object-track gap of \cref{sec:faithfulness}) a difference in flip rates. Chain GRPO uses outcome and format rewards; $^{\dagger}$\,mean\,$\pm$\,s.d.\ over three seeds.}
\label{tab:clevrer_main}
\input{tables/tab_clevrer_main}
\end{table*}

\section{Mixed-data arms on CLEVRER}
\label{sec:supp_mixed}
\cref{tab:clevrer_mixed} reports every arm of the mixed experiment on CLEVRER. Three observations. Mixing the two chain styles costs CLEVRER accuracy: the mixed chain SFT drops from 0.902 to 0.648 on explanatory questions (0.697 under a second seed, with the same blind gap of 0.27, IoU 0.65 against 0.66 and track gap 0.53 against 0.49), and chain GRPO recovers part of it (0.756). The plain mix trained on the same 4,400 CLEVRER questions with bare answers reaches 0.970--0.973 on explanatory questions, the highest of any model, and 0.906 over all items against 0.739, so the cost comes from the chains, not the data. The NExT-QA chain models do not transfer to CLEVRER (0.533 overall, blind gap 0.06, IoU 0.20), so synthetic chains are necessary for synthetic grounding and the mix is the only arm grounded on both benchmarks. Grounding of the mixed chains on CLEVRER stays at the level of the CLEVRER chain models (IoU 0.65--0.66). On CausalVQA no arm moves (\cref{sec:supp_causalvqa}).

\begin{table*}[h]
\centering\tabsize
\caption{\textbf{CLEVRER accuracy of the mixed-data arms.} Columns as in \cref{tab:clevrer_main}. Mixing chain styles costs synthetic accuracy (explanatory 0.90 to 0.65), which chain GRPO partly recovers; the plain mix on the same questions is the most accurate model; NExT-QA chain models do not transfer to CLEVRER.}
\label{tab:clevrer_mixed}
\input{tables/tab_clevrer_mixed}
\end{table*}

\paragraph{Chain-free models depend on the same evidence.} \cref{tab:chainfree} gives the gold-evidence and object-track gaps of the models that write no chain, next to the chain models. The zero-shot base, outcome-only GRPO and the plain mixed arms all lose more answers when the annotated evidence is blacked out than under the control, by as much as the chain models. A positive \citegap on CLEVRER therefore shows that a chain points at evidence the answer depends on, not that the dependence is created by the chain.

\begin{table*}[h]
\centering\tabsize
\caption{\textbf{Gold-evidence and object-track gaps on CLEVRER}, chain-free models above the rule and chain models below (95\% intervals).}
\label{tab:chainfree}
\input{tables/tab_chainfree}
\end{table*}

\section{Outcome-only GRPO does not collapse}
\label{sec:supp_collapse}
Under both training mixes (\cref{tab:clevrer_main}), GRPO with outcome and format rewards raises accuracy on every CLEVRER type (explanatory 0.70 to 0.94--0.96 per option), and the gain is video-dependent: the blind gap grows from 0.28 to 0.47--0.50 and the track gap holds or grows. Wu~\etal~\cite{wu2026beyond} report a 10-point drop for the same base model under outcome-only GRPO. Our protocol differs from theirs (LoRA with a KL penalty to the base, a different initialisation and prompt), so this is a non-replication under different conditions, not a refutation, and we did not test which difference matters. Their released checkpoint scores 0.85 on counterfactual and 0.82 on explanatory questions in our suite. Chain GRPO does lower the \citegap of the mixed model, from 0.089 to 0.042 (\cref{tab:cite_clevrer}), but the control flip rate halves with it and the ratio of the two is unchanged (2.1 and 2.2): the RL-trained model is more robust to occlusion in general, not less reliant on the cited box in particular.

\section{Process rewards and credit assignment}
\label{sec:supp_stage2}
\paragraph{Rewards.} The general chain reward is
\begin{equation}
R = R_{\text{out}} + 0.5\,R_{\text{fmt}} + w\,(R_{\text{gnd}} + R_{\text{nec}} + R_{\text{proc}}),
\label{eq:reward}
\end{equation}
where $R_{\text{out}}\in\{0,1\}$ is answer correctness and $R_{\text{fmt}}\in\{0,1\}$ whether the chain parses. The \emph{chain GRPO} arm of the main text sets $w=0$; the \emph{CSR} arm (chain-structured rewards) sets $w=0.5$ and adds three process terms. $R_{\text{gnd}}$ is the mean spatiotemporal IoU between each predicted step's box and its matched reference step, size-binned as in Ground-R1~\cite{cao2025groundr1}. $R_{\text{nec}}$ scores step necessity under a frozen verifier $p_\phi$ (the SFT checkpoint): with $\Delta_i = p_\phi(a^\ast\mid v,q,c) - p_\phi(a^\ast\mid v,q,c_{-i})$ and $c_{-i}$ the chain without step $i$ and its dependency closure,
\begin{equation}
R_{\text{nec}} = \tfrac{1}{N}\textstyle\sum_i \mathbf{1}[\Delta_i > \delta] \;-\; \gamma\,\tfrac{1}{N}\,|\{i : \Delta_i \le \varepsilon\}|,
\label{eq:nec}
\end{equation}
with $\delta=0.1$, $\varepsilon=0.02$, $\gamma=0.5$. $R_{\text{proc}}$ is H-GRPO's Hungarian match score~\cite{peh2026hgrpo} between predicted and reference steps (Sentence-BERT~\cite{reimers2019sbert} similarity times spatiotemporal IoU) minus a penalty for matched steps whose order reverses a reference dependency edge. Step-level credit assignment distributes the process reward over the tokens of each step, at weight 0.3 or 1.0.

\paragraph{Results.} \cref{tab:stage2} lists every RL-on-chains arm trained from the same SFT checkpoint on CLEVRER. CSR reaches 0.867\,$\pm$\,0.004 per option and box IoU 0.667\,$\pm$\,0.001 over two seeds, against 0.870\,$\pm$\,0.001 and 0.660\,$\pm$\,0.006 for chain GRPO over three: no difference in accuracy, and a grounding difference of about one between-seed standard deviation. The seed matters for how this reads. Paired over items at seed 0, CSR changes IoU by $+0.017$ \CI{+0.013}{+0.021} relative to chain GRPO, which looks like the process rewards protecting grounding; but a second chain GRPO seed differs from the first by $+0.015$ \CI{+0.010}{+0.019} on the same paired statistic, so an interval that treats the seed as fixed overstates the certainty of the effect. Dropping the necessity term gives IoU 0.637 (one seed). Step-level credit at weight 1.0 lowered the training reward, raised KL tenfold and held the gradient norm at the clip, and it has the lowest accuracy and IoU of the arms (0.842 and 0.607); at weight 0.3 training was stable and the model matched CSR in accuracy with IoU 0.658 and the lowest track gap of the chain models. We do not observe a consistent effect of the tested process rewards at this budget and seed count, which is two to three seeds and 1,000 steps and so cannot rule out a small one.

\begin{table*}[h]
\centering\tabsize
\caption{\textbf{RL-on-chains arms on CLEVRER} (2,537 per-option items). Format = share of parseable chains.}
\label{tab:stage2}
\input{tables/tab_stage2}
\end{table*}

\section{Second base: InternVL3-2B}
\label{sec:supp_internvl}
\cref{tab:internvl} reports the synthetic pipeline on InternVL3-2B (language model Qwen2.5-1.5B, vision encoder InternViT-300M). Two findings transfer. Outcome-only GRPO does not collapse on this base either: explanatory and predictive accuracy rise by 7 points and the blind gap grows. The chain pipeline produces well-grounded steps (IoU 0.74 after SFT, 0.67--0.73 after chain GRPO, 99\% parseable), and the chain GRPO model has a positive \citegap (\cref{tab:cite_clevrer}). Accuracy parity does not transfer: the InternVL chain models stay below the base on explanatory questions (0.69 for chain SFT and 0.76--0.81 for chain GRPO, against 0.86), whereas on Qwen they matched outcome-only GRPO. Learning rates: RL at the Qwen rate ($10^{-5}$) was unstable for the outcome-only arm (KL 11--58, falling reward) and is not reported; the chain GRPO run at that rate completed with intact accuracy (0.803) and track gap (0.55) and is the model in the InternVL row of \cref{tab:cite_clevrer}; \cref{tab:internvl} lists both it and the $3\times10^{-6}$ run.

\begin{table*}[h]
\centering\tabsize
\caption{\textbf{InternVL3-2B on CLEVRER.} Columns as in \cref{tab:clevrer_main}.}
\label{tab:internvl}
\input{tables/tab_internvl}
\end{table*}

\section{CausalVQA}
\label{sec:supp_causalvqa}
\cref{tab:causalvqa} reports the debug split of CausalVQA (200 choice sets; a question is \emph{paired}-correct when both of its choice sets are correct). Two effects stack. Every CLEVRER-trained model loses accuracy relative to the base under the letters prompt (outcome-only GRPO $-6.5$ points, the released VideoThinker-R1-3B $-10.5$). The per-option format then removes most of what remains, for the untrained base as much as for any trained model (0.580 to 0.115 set accuracy), so every chain arm sits at or below that floor. NExT-QA chains, which are observations of everyday scenes, do not transfer to CausalVQA's egocentric anticipation and planning questions.

\begin{table*}[h]
\centering\tabsize
\caption{\textbf{CausalVQA debug split.} Letters-prompt rows answer one multiple-choice prompt per set; the others answer one yes/no item per option.}
\label{tab:causalvqa}
\input{tables/tab_causalvqa}
\end{table*}

\section{Training, evaluation and compute}
\label{sec:supp_training}
\paragraph{Chains.} In the CLEVRER generator only entity-producing program steps become triplets (scalar operations such as count, exist, attribute query and negation fold into the final answer, so no step restates it), each with a role: \emph{resolve} (a unique object or event), \emph{enumerate} (a set consumed by a count or existence test) or \emph{observe} (a candidate collision from a choice program). Evidence boxes come from the released derender proposals; dropped chains are mostly those of objects visible for at most three frames. Chains have 1--3 steps for descriptive, 3--4 for predictive and counterfactual, and 4--7 for explanatory questions. Three conventions were calibrated against the ground-truth answers: a motion threshold on annotated speed and the end-frame index, on the first 1,000 training videos, and the treatment of collisions annotated while an object is off-screen, fixed after a full scan found 17 mismatches (such a collision is registered at the first frame where both objects are visible).

\setcounter{dbltopnumber}{3}\makeatletter\setlength{\@dblfptop}{0pt}\makeatother%
\paragraph{SFT.} LoRA rank 64, $\alpha=128$, dropout 0.05 on the q, k, v, o, gate, up and down projections; learning rate $10^{-4}$ with 30 warm-up steps and cosine decay; effective batch 8 by gradient accumulation; one epoch; bridge weight $\lambda=0.2$ on 50\% of examples; bfloat16 with gradient checkpointing. A seed fixes the data order and the choice of corrupted step. The 8,800 CLEVRER chains take 1,100 optimizer steps and the mixed set of 12,400 chain items 1,550. The bridge term is not ablated.

\paragraph{GRPO.} Group size 8, one prompt per step drawn from the 1,500-item RL pool, temperature 1.0, up to 640 new tokens (the longest SFT chain is 549 tokens), learning rate $10^{-5}$ ($3\times10^{-6}$ and $10^{-5}$ for InternVL, \cref{sec:supp_internvl}), KL coefficient 0.04 to the SFT model (the SFT adapter is merged and the reference is the merged weights with the RL adapter disabled), log-ratio clip 5, gradient clip 1.0, 1,000 steps. Outcome-only GRPO from the base uses the same settings for 1,500 steps. Reward weights follow \cref{eq:reward}; the necessity verifier is the SFT checkpoint evaluated on 8 of the 16 frames, with $\delta=0.1$, $\varepsilon=0.02$ and $\gamma=0.5$.

\paragraph{Pseudo-labels.} Teacher Qwen2.5-VL-7B-Instruct with 16 frames; Grounding DINO tiny at threshold 0.35; the detector box is the step's evidence; the frame index maps to the 128-position grid with a $\pm4$ span; one ``yes'' and one sampled ``no'' item per chain. Chains were written for 5,283 questions, and the 8,000 training items (4,000 of each answer) are a balanced sample of their items.

\paragraph{Evaluation.} 16 uniformly sampled frames decoded in a stream; greedy decoding; 640 new tokens for chain models, 64 for chain-free models, and 384 for published checkpoints with their own prompts (768 for VideoRFT-3B, which truncates); batch 8. The chain prompt is in \cref{sec:supp_prompt}. Bootstrap intervals use 2,000 paired resamples over questions (all per-option items of a question move together), seeded, so the reported digits are reproducible; item-level resampling gives the same intervals to two decimals in every case we checked. The splits by gold answer and by citation hit use the same procedure on the corresponding subsets.

\paragraph{Compute.} Every run used a single A100 80GB; nothing was trained or evaluated on more than one GPU. Training costs per run: chain SFT 2.3\,h, plain SFT 1.0\,h, chain GRPO (1,000 steps) 2.8\,h, CSR with the necessity verifier about 4\,h, outcome-only GRPO (1,500 steps) 4\,h. The 25 training runs behind the paper total roughly 70 GPU-hours, plus about 3\,h to pseudo-label the NExT-QA chains with the teacher and the detector. Evaluation dominates: one condition is a full pass over 2,537 or 4,995 items (0.5\,h for a chain-free model, 2.0--2.8\,h for a chain model; 2.2--2.3\,h for an entity-track condition, which covers the roughly 4,100 items that are correct under the plain input), and the 268 conditions reported here sum to 318 GPU-hours of recorded wall-clock. The Grounding DINO passes for the citation-grounding check and for the entity tracks (2,746 video--phrase pairs on 16 frames each) ran on the same GPU and are not included in that total. All statistics, including the bootstrap, the splits and the control replay, run on CPU in minutes.

\section{Prompt}
\label{sec:supp_prompt}
\begin{quote}\footnotesize\ttfamily
<question>\\
Option A: <option text>\\
Is this option correct? Answer yes or no.\\
You answer questions about a video by reasoning in grounded steps. Inside <think></think>, write one step per line as [n] Q: <sub-question> | A: <intermediate answer> | E: t=<start>-<end> box=(x0,y0,x1,y1) | deps=<comma-separated earlier step numbers>. t is a frame range of the video and the box is the evidence region in normalized coordinates. Then give the final answer inside <answer></answer>.
\end{quote}

%% file: tables/tab_pooled.tex
\begin{tabular}{@{}l c c c c c@{}}
\toprule
Test & $n$ & Flip [95\% CI] & Ctrl. & Ratio & Gap [95\% CI] \\
\midrule
\emph{CLEVRER, five chain models} &  &  &  &  &  \\
Self-citation (cited boxes), all items & 9403 & 0.097\,{\scriptsize[0.090,0.105]} & 0.045 & 2.2 & 0.053\,{\scriptsize[0.045,0.060]} \\
Self-citation (cited boxes), correct options & 4596 & 0.090\,{\scriptsize[0.079,0.102]} & 0.040 & 2.2 & 0.050\,{\scriptsize[0.040,0.059]} \\
Gold evidence boxes, all items & 9433 & 0.144\,{\scriptsize[0.133,0.155]} & 0.068 & 2.1 & 0.076\,{\scriptsize[0.067,0.085]} \\
Gold evidence boxes, correct options & 4613 & 0.147\,{\scriptsize[0.130,0.165]} & 0.073 & 2.0 & 0.074\,{\scriptsize[0.061,0.087]} \\
Object track (gold objects, all frames), all items & 8922 & 0.563\,{\scriptsize[0.549,0.578]} & 0.067 & 8.3 & 0.495\,{\scriptsize[0.479,0.511]} \\
Object track (gold objects, all frames), correct options & 4445 & 0.603\,{\scriptsize[0.574,0.630]} & 0.048 & 12.5 & 0.555\,{\scriptsize[0.527,0.583]} \\
\midrule
\emph{NExT-QA, five models trained on NExT-QA chains} &  &  &  &  &  \\
Self-citation (cited boxes), all items & 20666 & 0.023\,{\scriptsize[0.021,0.026]} & 0.021 & 1.1 & 0.003\,{\scriptsize[0.001,0.005]} \\
Self-citation (cited boxes), correct options & 3988 & 0.036\,{\scriptsize[0.029,0.043]} & 0.029 & 1.2 & 0.007\,{\scriptsize[0.002,0.012]} \\
Entity track (named entities, all frames; two models), all items & 8227 & 0.102\,{\scriptsize[0.093,0.111]} & 0.055 & 1.9 & 0.047\,{\scriptsize[0.039,0.056]} \\
Entity track (named entities, all frames; two models), correct options & 1556 & 0.180\,{\scriptsize[0.159,0.202]} & 0.076 & 2.4 & 0.103\,{\scriptsize[0.081,0.127]} \\
\bottomrule
\end{tabular}

%% file: tables/tab_vd.tex
\begin{tabular}{@{}l c c cc c c@{}}
\toprule
Test & Blind flip & $n$ & Flip & Ctrl. & Gap [95\% CI] & Gap / blind flip \\
\midrule
\emph{CLEVRER, five chain models} &  &  &  &  &  &  \\
Self-citation (cited boxes) & 0.48 & 4486 & 0.124 & 0.057 & 0.067\,{\scriptsize[0.056,0.078]} & 0.11 \\
Gold evidence boxes & 0.48 & 4502 & 0.193 & 0.090 & 0.103\,{\scriptsize[0.089,0.117]} & 0.16 \\
Object track (gold objects, all frames) & 0.49 & 4338 & 0.717 & 0.095 & 0.622\,{\scriptsize[0.602,0.642]} & 1.02 \\
\midrule
\emph{NExT-QA, five models trained with NExT-QA chains} &  &  &  &  &  &  \\
Self-citation (cited boxes) & 0.29 & 5926 & 0.045 & 0.039 & 0.006\,{\scriptsize[0.001,0.011]} & 0.01 \\
Entity track (two models) & 0.27 & 2259 & 0.225 & 0.129 & 0.096\,{\scriptsize[0.076,0.118]} & 0.17 \\
\bottomrule
\end{tabular}

%% file: tables/tab_cite_bytype.tex
\begin{tabular}{l cccc}
\toprule
Model & Descr. / Causal & Expl. / Temporal & Pred. / Descr. & CF \\
\midrule
CLEVRER chain GRPO & 0.035\,{\scriptsize[0.004,0.068]} & 0.042\,{\scriptsize[0.019,0.067]} & 0.082\,{\scriptsize[0.047,0.118]} & 0.050\,{\scriptsize[0.032,0.068]} \\
Mixed chain SFT & 0.019\,{\scriptsize[$-$0.009,0.047]} & 0.156\,{\scriptsize[0.112,0.200]} & 0.085\,{\scriptsize[0.055,0.116]} & 0.054\,{\scriptsize[0.030,0.079]} \\
Mixed chain GRPO & 0.000\,{\scriptsize[$-$0.034,0.032]} & 0.050\,{\scriptsize[0.026,0.076]} & 0.045\,{\scriptsize[0.018,0.073]} & 0.047\,{\scriptsize[0.028,0.068]} \\
NExT-QA chain GRPO & 0.129\,{\scriptsize[0.054,0.207]} & 0.011\,{\scriptsize[$-$0.009,0.031]} & 0.027\,{\scriptsize[0.009,0.047]} & 0.026\,{\scriptsize[0.002,0.048]} \\
InternVL3-2B chain GRPO & 0.059\,{\scriptsize[0.019,0.100]} & 0.025\,{\scriptsize[0.010,0.039]} & 0.108\,{\scriptsize[0.074,0.142]} & 0.031\,{\scriptsize[0.010,0.051]} \\
\midrule
CLEVRER chain GRPO & 0.017\,{\scriptsize[$-$0.002,0.034]} & 0.009\,{\scriptsize[$-$0.009,0.027]} & 0.019\,{\scriptsize[0.001,0.038]} & -- \\
Mixed chain SFT & 0.006\,{\scriptsize[$-$0.001,0.014]} & 0.002\,{\scriptsize[$-$0.006,0.010]} & 0.001\,{\scriptsize[$-$0.007,0.009]} & -- \\
Mixed chain SFT, seed 1 & 0.002\,{\scriptsize[$-$0.004,0.009]} & $-$0.001\,{\scriptsize[$-$0.008,0.007]} & 0.007\,{\scriptsize[$-$0.001,0.015]} & -- \\
Mixed chain GRPO & 0.008\,{\scriptsize[0.001,0.016]} & 0.004\,{\scriptsize[$-$0.003,0.012]} & $-$0.001\,{\scriptsize[$-$0.008,0.004]} & -- \\
NExT-QA chain SFT & 0.007\,{\scriptsize[$-$0.001,0.014]} & 0.001\,{\scriptsize[$-$0.009,0.011]} & 0.003\,{\scriptsize[$-$0.004,0.010]} & -- \\
NExT-QA chain GRPO & 0.000\,{\scriptsize[$-$0.006,0.006]} & 0.001\,{\scriptsize[$-$0.006,0.007]} & 0.001\,{\scriptsize[$-$0.004,0.007]} & -- \\
\bottomrule
\end{tabular}

%% file: tables/tab_nextqa.tex
\begin{tabular}{l cccccc}
\toprule
Model & Acc. (q) & Per-option & ``yes'' rate & Selection & Blind sel. & Blind gap \\
\midrule
Qwen2.5-VL-3B, letters prompt & 0.730 & -- & -- & -- & 0.437 & 0.29 \\
\quad + outcome-only GRPO (CLEVRER), letters prompt & 0.750 & -- & -- & -- & 0.437 & 0.31 \\
\midrule
Qwen2.5-VL-3B & 0.461 & 0.860 & 0.19 & 0.595 & 0.270 & 0.06 \\
\quad + outcome-only GRPO (CLEVRER) & 0.355 & 0.804 & 0.15 & 0.522 & 0.260 & 0.00 \\
CLEVRER chain SFT & 0.122 & 0.622 & 0.38 & 0.451 & 0.259 & 0.12 \\
CLEVRER chain GRPO & 0.194 & 0.682 & 0.41 & 0.470 & 0.272 & 0.12 \\
\midrule
Mixed plain SFT (bare answers) & 0.445 & 0.826 & 0.32 & 0.637 & 0.396 & 0.12 \\
Mixed plain SFT, seed 1 & 0.487 & 0.849 & 0.29 & 0.661 & 0.386 & 0.13 \\
\quad + outcome GRPO & 0.436 & 0.822 & 0.33 & 0.631 & 0.383 & 0.14 \\
Mixed chain SFT & 0.476 & 0.845 & 0.27 & 0.637 & 0.357 & 0.22 \\
Mixed chain SFT, seed 1 & 0.483 & 0.846 & 0.27 & 0.642 & 0.377 & 0.16 \\
Mixed chain GRPO & 0.463 & 0.837 & 0.29 & 0.640 & 0.358 & 0.25 \\
NExT-QA chain SFT & 0.458 & 0.837 & 0.30 & 0.641 & 0.368 & 0.19 \\
NExT-QA chain GRPO & 0.481 & 0.852 & 0.28 & 0.653 & 0.379 & 0.15 \\
\bottomrule
\end{tabular}

%% file: tables/tab_cite_nextqa.tex
\begin{tabular}{l cc c cc c}
\toprule
Model & Acc. & Blind gap & $n$ & Flip (cited) & Flip (random) & Cite gap [95\% CI] \\
\midrule
CLEVRER chain GRPO & 0.682 & 0.12 & 4770 & 0.085 & 0.070 & 0.015\,{\scriptsize[0.003,0.025]} \\
Mixed chain SFT & 0.845 & 0.22 & 4906 & 0.028 & 0.025 & 0.003\,{\scriptsize[$-$0.002,0.007]} \\
Mixed chain SFT, seed 1 & 0.846 & 0.16 & 4914 & 0.023 & 0.020 & 0.003\,{\scriptsize[$-$0.002,0.007]} \\
Mixed chain GRPO & 0.837 & 0.25 & 4912 & 0.022 & 0.018 & 0.004\,{\scriptsize[0.000,0.008]} \\
NExT-QA chain SFT & 0.837 & 0.19 & 4891 & 0.029 & 0.025 & 0.003\,{\scriptsize[$-$0.001,0.008]} \\
NExT-QA chain GRPO & 0.852 & 0.15 & 4889 & 0.016 & 0.015 & 0.001\,{\scriptsize[$-$0.003,0.004]} \\
\midrule
Control: plain SFT masked with mixed chain SFT's citations & 0.826 & 0.12 & 4906 & 0.014 & 0.016 & $-$0.002\,{\scriptsize[$-$0.005,0.001]} \\
\bottomrule
\end{tabular}

%% file: tables/tab_cite_track_nextqa.tex
\begin{tabular}{@{}l c cc c ccc c c@{}}
\toprule
Model & $n$ & Flip (tracks) & Flip (random) & Entity-track gap & Causal & Temporal & Descr. & Cite gap & Blind gap \\
\midrule
Mixed chain SFT & 4087 & 0.110 & 0.062 & 0.048\,{\scriptsize[0.038,0.059]} & 0.046 & 0.036 & 0.061 & 0.003\,{\scriptsize[$-$0.002,0.007]} & 0.22 \\
NExT-QA chain GRPO & 4140 & 0.095 & 0.049 & 0.046\,{\scriptsize[0.037,0.055]} & 0.043 & 0.031 & 0.065 & 0.001\,{\scriptsize[$-$0.003,0.004]} & 0.15 \\
\bottomrule
\end{tabular}

%% file: tables/tab_grounding.tex
\begin{tabular}{l c c c c}
\toprule
Model & Det. & IoU & Hit@0.5 & Steps \\
\midrule
CLEVRER chain GRPO & 0.889 & 0.25 / 0.10 & 0.12 / 0.00 & 18 \\
Mixed chain SFT & 0.900 & 0.54 / 0.23 & 0.60 / 0.18 & 14274 \\
Mixed chain GRPO & 0.916 & 0.56 / 0.24 & 0.62 / 0.19 & 14395 \\
NExT-QA chain SFT & 0.893 & 0.54 / 0.22 & 0.60 / 0.17 & 14410 \\
NExT-QA chain GRPO & 0.908 & 0.55 / 0.23 & 0.62 / 0.18 & 14448 \\
\bottomrule
\end{tabular}

%% file: tables/tab_gold_split.tex
\begin{tabular}{@{}l cccc cccc@{}}
\toprule
Model & $n$ & Flip & Ctrl. & Gap (correct option) & $n$ & Flip & Ctrl. & Gap (wrong option) \\
\midrule
\emph{NExT-QA, self-citation test} &  &  &  &  &  &  &  &  \\
CLEVRER chain GRPO & 787 & 0.062 & 0.062 & 0.000\,{\scriptsize[$-$0.019,0.019]} & 2449 & 0.092 & 0.073 & 0.019\,{\scriptsize[0.007,0.032]} \\
Mixed chain SFT & 769 & 0.057 & 0.046 & 0.012\,{\scriptsize[$-$0.001,0.026]} & 3375 & 0.021 & 0.020 & 0.001\,{\scriptsize[$-$0.004,0.006]} \\
Mixed chain SFT, seed 1 & 776 & 0.044 & 0.041 & 0.003\,{\scriptsize[$-$0.010,0.015]} & 3384 & 0.018 & 0.015 & 0.003\,{\scriptsize[$-$0.001,0.008]} \\
Mixed chain GRPO & 817 & 0.027 & 0.017 & 0.010\,{\scriptsize[0.001,0.018]} & 3292 & 0.020 & 0.018 & 0.002\,{\scriptsize[$-$0.002,0.007]} \\
NExT-QA chain SFT & 822 & 0.033 & 0.027 & 0.006\,{\scriptsize[$-$0.005,0.017]} & 3269 & 0.028 & 0.025 & 0.003\,{\scriptsize[$-$0.003,0.009]} \\
NExT-QA chain GRPO & 804 & 0.019 & 0.014 & 0.005\,{\scriptsize[$-$0.004,0.014]} & 3358 & 0.015 & 0.016 & 0.000\,{\scriptsize[$-$0.004,0.004]} \\
\midrule
\emph{NExT-QA, entity-track test} &  &  &  &  &  &  &  &  \\
Mixed chain SFT & 758 & 0.194 & 0.096 & 0.098\,{\scriptsize[0.069,0.127]} & 3329 & 0.090 & 0.054 & 0.037\,{\scriptsize[0.026,0.048]} \\
NExT-QA chain GRPO & 798 & 0.167 & 0.058 & 0.109\,{\scriptsize[0.084,0.134]} & 3342 & 0.078 & 0.046 & 0.031\,{\scriptsize[0.023,0.040]} \\
\midrule
\emph{CLEVRER, self-citation test} &  &  &  &  &  &  &  &  \\
CLEVRER chain GRPO & 1006 & 0.090 & 0.037 & 0.054\,{\scriptsize[0.035,0.074]} & 1000 & 0.084 & 0.033 & 0.051\,{\scriptsize[0.033,0.070]} \\
Mixed chain SFT & 888 & 0.175 & 0.087 & 0.088\,{\scriptsize[0.061,0.116]} & 794 & 0.184 & 0.078 & 0.106\,{\scriptsize[0.077,0.135]} \\
Mixed chain GRPO & 1094 & 0.049 & 0.020 & 0.029\,{\scriptsize[0.016,0.044]} & 664 & 0.136 & 0.062 & 0.074\,{\scriptsize[0.047,0.101]} \\
NExT-QA chain GRPO & 707 & 0.052 & 0.031 & 0.021\,{\scriptsize[0.004,0.037]} & 563 & 0.064 & 0.037 & 0.027\,{\scriptsize[0.008,0.046]} \\
InternVL3-2B chain GRPO & 901 & 0.085 & 0.030 & 0.055\,{\scriptsize[0.039,0.073]} & 962 & 0.073 & 0.038 & 0.034\,{\scriptsize[0.018,0.052]} \\
\bottomrule
\end{tabular}

%% file: tables/tab_hit_split.tex
\begin{tabular}{@{}l ccc ccc@{}}
\toprule
Model & Items & Cite gap & Entity-track gap & Items & Cite gap & Entity-track gap \\
\midrule
Mixed chain SFT & 1612 & 0.011\,{\scriptsize[0.003,0.019]} & 0.042\,{\scriptsize[0.025,0.059]} & 3294 & $-$0.001\,{\scriptsize[$-$0.007,0.004]} & 0.051\,{\scriptsize[0.037,0.065]} \\
Mixed chain SFT, seed 1 & 1446 & 0.005\,{\scriptsize[$-$0.004,0.014]} & -- & 3468 & 0.002\,{\scriptsize[$-$0.002,0.007]} & -- \\
Mixed chain GRPO & 1687 & $-$0.001\,{\scriptsize[$-$0.009,0.007]} & -- & 3225 & 0.006\,{\scriptsize[0.002,0.011]} & -- \\
NExT-QA chain SFT & 1440 & $-$0.001\,{\scriptsize[$-$0.010,0.008]} & -- & 3451 & 0.005\,{\scriptsize[$-$0.001,0.011]} & -- \\
NExT-QA chain GRPO & 1689 & 0.000\,{\scriptsize[$-$0.006,0.006]} & 0.049\,{\scriptsize[0.034,0.065]} & 3200 & 0.001\,{\scriptsize[$-$0.003,0.005]} & 0.045\,{\scriptsize[0.034,0.056]} \\
\bottomrule
\end{tabular}

%% file: tables/tab_clevrer_strata.tex
\begin{tabular}{@{}l cc cc cc cc@{}}
\toprule
Model & $n$ & Gap, all boxes hit & $n$ & Gap, a box misses & Ctrl. & Gap, ctrl.\ on object & Ctrl. & Gap, ctrl.\ on background \\
\midrule
CLEVRER chain GRPO & 974 & 0.061\,{\scriptsize[0.042,0.079]} & 1229 & 0.044\,{\scriptsize[0.026,0.062]} & 0.050 & 0.044\,{\scriptsize[0.027,0.062]} & 0.014 & 0.062\,{\scriptsize[0.046,0.079]} \\
Mixed chain SFT & 878 & 0.077\,{\scriptsize[0.056,0.100]} & 986 & 0.098\,{\scriptsize[0.070,0.126]} & 0.105 & 0.083\,{\scriptsize[0.058,0.108]} & 0.035 & 0.097\,{\scriptsize[0.072,0.122]} \\
Mixed chain GRPO & 952 & 0.045\,{\scriptsize[0.029,0.062]} & 993 & 0.039\,{\scriptsize[0.020,0.058]} & 0.049 & 0.040\,{\scriptsize[0.022,0.060]} & 0.022 & 0.044\,{\scriptsize[0.028,0.060]} \\
NExT-QA chain GRPO & 11 & 0.000\,{\scriptsize[$-$0.188,0.333]} & 90 & 0.056\,{\scriptsize[0.000,0.112]} & 0.058 & 0.007\,{\scriptsize[$-$0.017,0.033]} & 0.029 & 0.041\,{\scriptsize[0.026,0.056]} \\
InternVL3-2B chain GRPO & 950 & 0.065\,{\scriptsize[0.047,0.086]} & 1084 & 0.030\,{\scriptsize[0.016,0.045]} & 0.048 & 0.034\,{\scriptsize[0.015,0.051]} & 0.026 & 0.060\,{\scriptsize[0.044,0.077]} \\
\bottomrule
\end{tabular}

%% file: tables/tab_symmetric.tex
\begin{tabular}{@{}l cc cc cc@{}}
\toprule
Model & Net acc., mask & Net acc., ctrl. & W$\to$C, mask & W$\to$C, ctrl. & Q.\ flip, mask & Q.\ flip, ctrl. \\
\midrule
\emph{NExT-QA, self-citation test} &  &  &  &  &  &  \\
Mixed chain SFT & $+$0.001 & $-$0.001 & 0.160 & 0.131 & 0.095 & 0.101 \\
Mixed chain SFT, seed 1 & $-$0.001 & $-$0.003 & 0.122 & 0.094 & 0.074 & 0.070 \\
Mixed chain GRPO & $-$0.001 & $-$0.002 & 0.106 & 0.082 & 0.073 & 0.057 \\
NExT-QA chain SFT & $-$0.004 & $-$0.003 & 0.124 & 0.113 & 0.097 & 0.112 \\
NExT-QA chain GRPO & $-$0.001 & $-$0.001 & 0.087 & 0.080 & 0.062 & 0.054 \\
\midrule
\emph{NExT-QA, entity-track test} &  &  &  &  &  &  \\
Mixed chain SFT & $-$0.110 & $-$0.062 & -- & -- & 0.359 & 0.220 \\
NExT-QA chain GRPO & $-$0.095 & $-$0.049 & -- & -- & 0.348 & 0.188 \\
\midrule
\emph{CLEVRER, self-citation test} &  &  &  &  &  &  \\
CLEVRER chain GRPO & $-$0.047 & $-$0.016 & 0.225 & 0.120 & -- & -- \\
Mixed chain SFT & $-$0.067 & $-$0.020 & 0.215 & 0.145 & -- & -- \\
Mixed chain GRPO & $-$0.038 & $-$0.013 & 0.096 & 0.063 & -- & -- \\
NExT-QA chain GRPO & $-$0.009 & $-$0.006 & 0.060 & 0.031 & -- & -- \\
InternVL3-2B chain GRPO & $-$0.041 & $-$0.014 & 0.135 & 0.080 & -- & -- \\
\bottomrule
\end{tabular}

%% file: tables/tab_dose.tex
\begin{tabular}{@{}l cccc@{}}
\toprule
Model &  &  &  &  \\
\midrule
\emph{Cite gap by share of the entity's frames that are cited} & $<0.15$ & 0.15--0.3 & $\ge0.3$ &  \\
Mixed chain SFT & 0.000\,{\scriptsize[$-$0.006,0.006]} & $-$0.001\,{\scriptsize[$-$0.008,0.005]} & 0.023\,{\scriptsize[0.010,0.037]} &  \\
Mixed chain SFT, seed 1 & 0.002\,{\scriptsize[$-$0.004,0.009]} & 0.003\,{\scriptsize[$-$0.003,0.010]} & $-$0.002\,{\scriptsize[$-$0.013,0.007]} &  \\
Mixed chain GRPO & 0.001\,{\scriptsize[$-$0.004,0.005]} & 0.006\,{\scriptsize[$-$0.001,0.015]} & 0.004\,{\scriptsize[$-$0.006,0.016]} &  \\
NExT-QA chain SFT & $-$0.002\,{\scriptsize[$-$0.009,0.005]} & 0.008\,{\scriptsize[0.001,0.016]} & 0.002\,{\scriptsize[$-$0.009,0.012]} &  \\
NExT-QA chain GRPO & $-$0.008\,{\scriptsize[$-$0.017,0.000]} & 0.004\,{\scriptsize[0.000,0.008]} & 0.002\,{\scriptsize[$-$0.006,0.008]} &  \\
\midrule
\emph{Entity-track gap by frames on which the entity is detected} & 1--8 & 9--13 & 14--15 & 16 \\
Mixed chain SFT & 0.000\,{\scriptsize[$-$0.036,0.036]} & 0.039\,{\scriptsize[0.017,0.064]} & 0.054\,{\scriptsize[0.032,0.076]} & 0.050\,{\scriptsize[0.035,0.065]} \\
NExT-QA chain GRPO & 0.038\,{\scriptsize[0.000,0.096]} & 0.031\,{\scriptsize[0.014,0.047]} & 0.041\,{\scriptsize[0.025,0.059]} & 0.054\,{\scriptsize[0.042,0.068]} \\
\bottomrule
\end{tabular}

%% file: tables/tab_control.tex
\begin{tabular}{@{}l c cc c cc@{}}
\toprule
Model & Frames & Covered (\%) & Untouched (\%) & $n$ & Flip (cited) [95\% CI] & Cite gap, untouched \\
\midrule
\emph{NExT-QA} &  &  &  &  &  &  \\
CLEVRER chain GRPO & 1.3 & 22 & 44 & 1431 & 0.085\,{\scriptsize[0.070,0.099]} & 0.016\,{\scriptsize[0.001,0.032]} \\
Mixed chain SFT & 2.0 & 40 & 18 & 743 & 0.020\,{\scriptsize[0.010,0.032]} & 0.003\,{\scriptsize[$-$0.008,0.014]} \\
Mixed chain SFT, seed 1 & 2.0 & 39 & 20 & 798 & 0.019\,{\scriptsize[0.010,0.029]} & 0.005\,{\scriptsize[$-$0.004,0.015]} \\
Mixed chain GRPO & 2.0 & 40 & 21 & 846 & 0.015\,{\scriptsize[0.007,0.025]} & 0.005\,{\scriptsize[$-$0.005,0.014]} \\
NExT-QA chain SFT & 2.0 & 38 & 17 & 701 & 0.026\,{\scriptsize[0.015,0.038]} & 0.000\,{\scriptsize[$-$0.014,0.014]} \\
NExT-QA chain GRPO & 2.0 & 38 & 17 & 671 & 0.019\,{\scriptsize[0.010,0.031]} & 0.006\,{\scriptsize[$-$0.004,0.017]} \\
\midrule
\emph{CLEVRER} &  &  &  &  &  &  \\
CLEVRER chain GRPO & 1.4 & 9 & 41 & 896 & 0.088\,{\scriptsize[0.070,0.107]} & 0.071\,{\scriptsize[0.054,0.089]} \\
Mixed chain SFT & 1.5 & 10 & 40 & 754 & 0.121\,{\scriptsize[0.098,0.146]} & 0.093\,{\scriptsize[0.070,0.116]} \\
Mixed chain GRPO & 1.4 & 8 & 49 & 995 & 0.067\,{\scriptsize[0.053,0.084]} & 0.050\,{\scriptsize[0.035,0.066]} \\
NExT-QA chain GRPO & 2.0 & 6 & 66 & 896 & 0.068\,{\scriptsize[0.052,0.085]} & 0.036\,{\scriptsize[0.021,0.052]} \\
InternVL3-2B chain GRPO & 1.2 & 8 & 49 & 999 & 0.083\,{\scriptsize[0.066,0.101]} & 0.052\,{\scriptsize[0.036,0.070]} \\
\bottomrule
\end{tabular}

%% file: tables/tab_text_split.tex
\begin{tabular}{@{}l cccc cccc@{}}
\toprule
Model & Items & Cite gap & E.-t. & E.-t., corr. & Items & Cite gap & E.-t. & E.-t., corr. \\
\midrule
Mixed chain SFT & 1943 & 0.008\,{\scriptsize[0.000,0.015]} & 0.052\,{\scriptsize[0.036,0.069]} & 0.081\,{\scriptsize[0.045,0.120]} & 1852 & 0.002\,{\scriptsize[$-$0.004,0.009]} & 0.046\,{\scriptsize[0.030,0.063]} & 0.115\,{\scriptsize[0.063,0.171]} \\
Mixed chain SFT, seed 1 & 1921 & 0.001\,{\scriptsize[$-$0.006,0.007]} & -- & -- & 1810 & 0.003\,{\scriptsize[$-$0.004,0.010]} & -- & -- \\
Mixed chain GRPO & 2231 & 0.006\,{\scriptsize[$-$0.001,0.014]} & -- & -- & 1703 & $-$0.001\,{\scriptsize[$-$0.007,0.005]} & -- & -- \\
NExT-QA chain SFT & 1957 & 0.005\,{\scriptsize[$-$0.004,0.014]} & -- & -- & 1867 & 0.002\,{\scriptsize[$-$0.004,0.010]} & -- & -- \\
NExT-QA chain GRPO & 2206 & 0.000\,{\scriptsize[$-$0.006,0.006]} & 0.038\,{\scriptsize[0.026,0.051]} & 0.086\,{\scriptsize[0.047,0.127]} & 1783 & 0.003\,{\scriptsize[$-$0.002,0.008]} & 0.049\,{\scriptsize[0.034,0.063]} & 0.120\,{\scriptsize[0.079,0.162]} \\
\bottomrule
\end{tabular}

%% file: tables/tab_entity_control.tex
\begin{tabular}{@{}ll c cc c c@{}}
\toprule
Model & Items & $n$ & Flip & Ctrl. & Entity-track gap & Gap, correct options \\
\midrule
Mixed chain SFT & Control mostly on another entity & 749 & 0.132 & 0.095 & 0.037\,{\scriptsize[0.011,0.064]} & 0.070\,{\scriptsize[0.007,0.141]} \\
 & Mixed & 1297 & 0.108 & 0.056 & 0.052\,{\scriptsize[0.033,0.070]} & 0.085\,{\scriptsize[0.033,0.134]} \\
 & Control on background & 2041 & 0.102 & 0.053 & 0.049\,{\scriptsize[0.035,0.065]} & 0.116\,{\scriptsize[0.076,0.157]} \\
\midrule
NExT-QA chain GRPO & Control mostly on another entity & 768 & 0.099 & 0.055 & 0.044\,{\scriptsize[0.024,0.067]} & 0.111\,{\scriptsize[0.049,0.174]} \\
 & Mixed & 1314 & 0.093 & 0.047 & 0.046\,{\scriptsize[0.030,0.061]} & 0.124\,{\scriptsize[0.085,0.166]} \\
 & Control on background & 2058 & 0.095 & 0.047 & 0.048\,{\scriptsize[0.035,0.061]} & 0.099\,{\scriptsize[0.061,0.137]} \\
\bottomrule
\end{tabular}

%% file: tables/tab_recite.tex
\begin{tabular}{@{}l c cccc cccc@{}}
\toprule
Model & Masked & Same box & Same fr. & Other fr. & Same ent. & Same box & Same fr. & Other fr. & Same ent. \\
\midrule
\emph{NExT-QA} &  &  &  &  &  &  &  &  &  \\
CLEVRER chain GRPO & 1.4 & 0.49 & 0.33 & 0.18 & 0.41 & 0.57 & 0.27 & 0.16 & 0.59 \\
Mixed chain SFT & 3.9 & 0.14 & 0.70 & 0.16 & 0.70 & 0.52 & 0.35 & 0.13 & 0.75 \\
Mixed chain SFT, seed 1 & 3.5 & 0.16 & 0.70 & 0.14 & 0.69 & 0.53 & 0.35 & 0.11 & 0.74 \\
Mixed chain GRPO & 3.6 & 0.15 & 0.67 & 0.18 & 0.74 & 0.52 & 0.32 & 0.15 & 0.77 \\
NExT-QA chain SFT & 4.0 & 0.15 & 0.74 & 0.11 & 0.68 & 0.54 & 0.37 & 0.09 & 0.73 \\
NExT-QA chain GRPO & 4.1 & 0.17 & 0.70 & 0.13 & 0.71 & 0.55 & 0.35 & 0.10 & 0.76 \\
\midrule
\emph{CLEVRER} &  &  &  &  &  &  &  &  &  \\
CLEVRER chain GRPO & 1.9 & 0.44 & 0.29 & 0.27 & -- & 0.75 & 0.14 & 0.12 & -- \\
Mixed chain SFT & 2.4 & 0.48 & 0.32 & 0.20 & -- & 0.81 & 0.11 & 0.07 & -- \\
Mixed chain GRPO & 1.4 & 0.53 & 0.28 & 0.19 & -- & 0.83 & 0.09 & 0.08 & -- \\
NExT-QA chain GRPO & 0.4 & 0.45 & 0.43 & 0.12 & 0.80 & 0.77 & 0.14 & 0.09 & 0.86 \\
InternVL3-2B chain GRPO & 0.5 & 0.58 & 0.27 & 0.16 & -- & 0.83 & 0.11 & 0.06 & -- \\
\bottomrule
\end{tabular}

%% file: tables/tab_audit.tex
\begin{tabular}{@{}l c ccc@{}}
\toprule
Judgement & $n$ & Yes & No & Unsure \\
\midrule
Step: entity visible on cited frame & 299 & 0.82 & 0.09 & 0.08 \\
Step: box on the named entity & 299 & 0.70 & 0.18 & 0.12 \\
Step: relevant to the question & 299 & 0.95 & 0.04 & 0.01 \\
\midrule
Chain: steps suffice to answer & 100 & 0.89 & 0.08 & 0.03 \\
Chain: a step beyond locating & 100 & 0.08 & 0.92 & 0.00 \\
\bottomrule
\end{tabular}

%% file: tables/tab_model_audit.tex
\begin{tabular}{@{}l ccc@{}}
\toprule
Judgement & Labels & SFT & GRPO \\
\midrule
Step: entity visible on cited frame & 0.82 \scriptsize{(299)} & 0.82 \scriptsize{(151)} & 0.79 \scriptsize{(150)} \\
Step: box on the named entity & 0.70 \scriptsize{(299)} & 0.52 \scriptsize{(151)} & 0.53 \scriptsize{(150)} \\
\quad where the entity is visible & 0.84 \scriptsize{(246)} & 0.63 \scriptsize{(124)} & 0.68 \scriptsize{(118)} \\
Step: relevant to the question & 0.95 \scriptsize{(299)} & 1.00 \scriptsize{(151)} & 1.00 \scriptsize{(150)} \\
\midrule
Chain: steps suffice to answer & 0.89 \scriptsize{(100)} & 0.52 \scriptsize{(50)} & 0.50 \scriptsize{(50)} \\
Chain: a step beyond locating & 0.08 \scriptsize{(100)} & 0.06 \scriptsize{(50)} & 0.04 \scriptsize{(50)} \\
Chain: steps support the verdict & -- & 0.62 \scriptsize{(50)} & 0.64 \scriptsize{(50)} \\
\bottomrule
\end{tabular}

%% file: tables/tab_clevrer_main.tex
\begin{tabular}{@{}l ccccc c cc@{}}
\toprule
Model & Descr. & Expl. & Pred. & CF & All (opt.) & Box IoU & Blind gap & Track gap \\
\midrule
Qwen2.5-VL-3B (zero-shot) & 0.720 & 0.702 & 0.764 & 0.712 & -- & -- & 0.28 & 0.56 \\
\quad + outcome-only GRPO, balanced & 0.776 & 0.939 & 0.900 & 0.820 & -- & -- & 0.47 & 0.59 \\
\quad + outcome-only GRPO, natural & 0.776 & 0.957 & 0.868 & 0.823 & -- & -- & 0.50 & 0.63 \\
\midrule
VideoRFT-3B~\cite{wang2025videorft} & 0.752 & 0.542 & 0.752 & 0.626 & -- & -- & 0.27 & 0.49 \\
VideoThinker-R1-3B~\cite{wu2026beyond} & 0.748 & 0.824 & 0.836 & 0.848 & -- & -- & 0.49 & 0.56 \\
Video-R1-7B~\cite{feng2025videor1} & 0.780 & 0.823 & 0.686 & 0.718 & -- & -- & 0.35 & 0.58 \\
\midrule
CLEVRER chain SFT (ours) & 0.888 & 0.902 & 0.820 & 0.805 & 0.851 & 0.679 & 0.39 & 0.51 \\
CLEVRER chain GRPO (ours)$^{\dagger}$ & 0.911{\scriptsize\,$\pm$\,0.012} & 0.949{\scriptsize\,$\pm$\,0.003} & 0.833{\scriptsize\,$\pm$\,0.003} & 0.799{\scriptsize\,$\pm$\,0.002} & 0.870{\scriptsize\,$\pm$\,0.001} & 0.660{\scriptsize\,$\pm$\,0.006} & 0.38{\scriptsize\,$\pm$\,0.02} & 0.50{\scriptsize\,$\pm$\,0.02} \\
\bottomrule
\end{tabular}

%% file: tables/tab_clevrer_mixed.tex
\begin{tabular}{l ccccc c cc}
\toprule
Model & Descr. & Expl. & Pred. & CF & All (opt.) & Box IoU & Blind gap & Track gap \\
\midrule
CLEVRER chain SFT & 0.888 & 0.902 & 0.820 & 0.805 & 0.851 & 0.679 & 0.39 & 0.51 \\
CLEVRER chain GRPO & 0.924 & 0.947 & 0.832 & 0.801 & 0.871 & 0.654 & 0.37 & 0.50 \\
\midrule
Mixed plain SFT (bare answers) & 0.892 & 0.970 & 0.900 & 0.848 & 0.906 & -- & 0.39 & 0.48 \\
Mixed plain SFT, seed 1 & 0.864 & 0.973 & 0.880 & 0.856 & 0.903 & -- & 0.42 & 0.45 \\
\quad + outcome GRPO & 0.904 & 0.971 & 0.892 & 0.843 & 0.904 & -- & 0.39 & 0.49 \\
Mixed chain SFT & 0.860 & 0.648 & 0.798 & 0.763 & 0.739 & 0.658 & 0.27 & 0.49 \\
Mixed chain SFT, seed 1 & 0.856 & 0.697 & 0.800 & 0.770 & 0.758 & 0.650 & 0.27 & 0.53 \\
Mixed chain GRPO & 0.880 & 0.756 & 0.764 & 0.759 & 0.771 & 0.649 & 0.29 & 0.48 \\
NExT-QA chain SFT & 0.476 & 0.512 & 0.622 & 0.521 & 0.533 & 0.201 & 0.06 & 0.46 \\
NExT-QA chain GRPO & 0.464 & 0.485 & 0.674 & 0.529 & 0.535 & 0.204 & 0.08 & 0.44 \\
\bottomrule
\end{tabular}

%% file: tables/tab_chainfree.tex
\begin{tabular}{@{}l cc c c@{}}
\toprule
Model & Flip (evidence) & Flip (random) & Gold-evidence gap & Object-track gap \\
\midrule
Qwen2.5-VL-3B (zero-shot) & 0.133 & 0.069 & 0.063\,{\scriptsize[0.035,0.094]} & 0.560\,{\scriptsize[0.512,0.609]} \\
\quad + outcome-only GRPO, balanced & 0.204 & 0.085 & 0.119\,{\scriptsize[0.092,0.149]} & 0.589\,{\scriptsize[0.547,0.628]} \\
\quad + outcome-only GRPO, natural & 0.191 & 0.093 & 0.097\,{\scriptsize[0.069,0.126]} & 0.627\,{\scriptsize[0.590,0.666]} \\
Mixed plain SFT & 0.107 & 0.040 & 0.067\,{\scriptsize[0.054,0.080]} & 0.482\,{\scriptsize[0.461,0.503]} \\
\quad + outcome GRPO & 0.108 & 0.037 & 0.071\,{\scriptsize[0.058,0.085]} & 0.488\,{\scriptsize[0.466,0.507]} \\
\midrule
CLEVRER chain SFT & 0.183 & 0.090 & 0.092\,{\scriptsize[0.071,0.114]} & 0.510\,{\scriptsize[0.484,0.534]} \\
CLEVRER chain GRPO & 0.126 & 0.050 & 0.076\,{\scriptsize[0.060,0.092]} & 0.497\,{\scriptsize[0.475,0.519]} \\
Mixed chain SFT & 0.188 & 0.094 & 0.094\,{\scriptsize[0.074,0.113]} & 0.487\,{\scriptsize[0.460,0.514]} \\
Mixed chain GRPO & 0.117 & 0.050 & 0.067\,{\scriptsize[0.052,0.082]} & 0.483\,{\scriptsize[0.458,0.507]} \\
\bottomrule
\end{tabular}

%% file: tables/tab_stage2.tex
\begin{tabular}{l c c c c cc}
\toprule
Model & Seeds & Acc. (opt.) & Box IoU & Format & Blind gap & Track gap \\
\midrule
CLEVRER chain SFT & 1 & 0.851 & 0.679 & 0.991 & 0.39 & 0.51 \\
CLEVRER chain GRPO (outcome + format) & 3 & 0.870{\scriptsize\,$\pm$\,0.001} & 0.660{\scriptsize\,$\pm$\,0.006} & 0.997 & 0.38 & 0.50 \\
\quad + CSR (outcome, format, grounding, necessity, process) & 2 & 0.867{\scriptsize\,$\pm$\,0.004} & 0.667{\scriptsize\,$\pm$\,0.001} & 0.998 & 0.37 & 0.49 \\
\quad + CSR without necessity & 1 & 0.866 & 0.637 & 0.996 & 0.40 & 0.51 \\
\quad + CSR, step-level credit 0.3 & 1 & 0.867 & 0.658 & 0.990 & 0.38 & 0.46 \\
\quad + CSR, step-level credit 1.0 & 1 & 0.842 & 0.607 & 0.962 & 0.38 & 0.50 \\
\bottomrule
\end{tabular}

%% file: tables/tab_internvl.tex
\begin{tabular}{l ccccc c cc}
\toprule
Model & Descr. & Expl. & Pred. & CF & All (opt.) & Box IoU & Blind gap & Track gap \\
\midrule
InternVL3-2B (zero-shot) & 0.836 & 0.855 & 0.826 & 0.764 & -- & -- & 0.42 & 0.51 \\
\quad + outcome-only GRPO & 0.856 & 0.927 & 0.896 & 0.770 & -- & -- & 0.50 & 0.56 \\
Chain SFT & 0.824 & 0.694 & 0.794 & 0.773 & 0.754 & 0.741 & 0.28 & 0.55 \\
\quad + chain GRPO, lr $3\times10^{-6}$ & 0.840 & 0.762 & 0.806 & 0.776 & 0.783 & 0.734 & 0.30 & 0.57 \\
\quad + chain GRPO, lr $10^{-5}$ (cite-test model) & 0.820 & 0.812 & 0.816 & 0.780 & 0.803 & 0.674 & 0.33 & 0.55 \\
\bottomrule
\end{tabular}

%% file: tables/tab_causalvqa.tex
\begin{tabular}{l cccc}
\toprule
Model & Set acc. & Paired & Per-option & Blind set acc. \\
\midrule
Qwen2.5-VL-3B, letters prompt & 0.580 & 0.430 & -- & 0.410 \\
\quad + outcome-only GRPO (CLEVRER), letters prompt & 0.515 & 0.350 & -- & 0.390 \\
VideoThinker-R1-3B~\cite{wu2026beyond}, letters prompt & 0.475 & 0.310 & -- & 0.340 \\
\midrule
Qwen2.5-VL-3B & 0.115 & 0.050 & 0.616 & 0.070 \\
\quad + outcome-only GRPO (CLEVRER) & 0.125 & 0.070 & 0.615 & 0.080 \\
CLEVRER chain SFT & 0.020 & 0.010 & 0.461 & 0.020 \\
CLEVRER chain GRPO & 0.050 & 0.020 & 0.479 & 0.010 \\
\quad + CSR & 0.040 & 0.020 & 0.426 & 0.015 \\
\midrule
Mixed plain SFT & 0.025 & 0.010 & 0.353 & 0.005 \\
\quad + outcome GRPO & 0.020 & 0.010 & 0.342 & 0.005 \\
Mixed chain SFT & 0.090 & 0.070 & 0.481 & 0.005 \\
Mixed chain GRPO & 0.040 & 0.030 & 0.382 & 0.000 \\
NExT-QA chain SFT & 0.030 & 0.020 & 0.378 & 0.010 \\
NExT-QA chain GRPO & 0.030 & 0.020 & 0.383 & 0.010 \\
\bottomrule
\end{tabular}